%% file: neurips_2025.tex
\documentclass{article}

    \usepackage[preprint]{neurips_2025}

\usepackage[utf8]{inputenc} % allow utf-8 input
\usepackage[T1]{fontenc}    % use 8-bit T1 fonts
\usepackage{hyperref}       % hyperlinks
\usepackage{url}            % simple URL typesetting
\usepackage{booktabs}       % professional-quality tables
\usepackage{amsfonts}       % blackboard math symbols
\usepackage{nicefrac}       % compact symbols for 1/2, etc.
\usepackage{microtype}      % microtypography
\usepackage{xcolor}   % colors
\usepackage{enumitem}
\definecolor{airforceblue}{rgb}{0.36, 0.54, 0.66}
\definecolor{aquamarine}{rgb}{0.5, 1.0, 0.83}
\definecolor{babyblue}{rgb}{0.54, 0.81, 0.94}
\definecolor{babyblueeyes}{rgb}{0.63, 0.79, 0.95}
\definecolor{blue(munsell)}{rgb}{0.0, 0.5, 0.69}
\definecolor{blue(ncs)}{rgb}{0.0, 0.53, 0.74}
\definecolor{ceruleanblue}{rgb}{0.16, 0.32, 0.75}
\definecolor{cobalt}{rgb}{0.0, 0.28, 0.67}
\definecolor{darkcerulean}{rgb}{0.03, 0.27, 0.49}
\definecolor{darkpowderblue}{rgb}{0.0, 0.2, 0.6}
\definecolor{denim}{rgb}{0.08, 0.38, 0.74}
\definecolor{egyptianblue}{rgb}{0.06, 0.2, 0.65}

\usepackage{algorithm}
\usepackage{algorithmic}

\usepackage{amsmath}
\usepackage{amssymb}
\usepackage{mathtools}
\usepackage{amsthm}
\usepackage{multirow}
\usepackage{subfigure}
\usepackage{array}
\usepackage{makecell}

\usepackage{wrapfig}
\usepackage{booktabs}
\usepackage{tabularx}
\usepackage{makecell}  % Optional: for better cell formatting
\newcommand\Tstrut{\rule{0pt}{0.335cm}}         % = `top' strut
\usepackage[capitalize,noabbrev]{cleveref}

\theoremstyle{plain}

\theoremstyle{definition}

\theoremstyle{remark}

\usepackage{mathtools}

\DeclarePairedDelimiter\floor{\lfloor}{\rfloor}

\usepackage{fixfoot}
\DeclareFixedFootnote{\repnote}{cf. $44.76_{\scriptscriptstyle \pm1.58}$ for a Random subset.} 

\title{Coreset Selection via LLM-based Concept Bottlenecks}

\author{Akshay Mehra\textsuperscript{1*}, Trisha Mittal\textsuperscript{1*}, Subhadra Gopalakrishnan\textsuperscript{2}, and Joshua Kimball\textsuperscript{1}\\
{\small \textsuperscript{*}Equal Contribution \textsuperscript{1}Dolby Laboratory \textsuperscript{2}Apple}\\ 
{\tt\small\{akshay.mehra, trisha.mittal, joshua.kimball\}@dolby.com}\\
}

\begin{document}

\maketitle

\input{sections/00-abstract}
\input{sections/01-introduction}

\input{sections/02-relatedwork}

\input{sections/03-preliminaries}
\input{sections/04-approach}
\input{sections/05-experiments}
\input{sections/06-conclusion}


\bibliography{main_paper}
\bibliographystyle{plain}

%%%%%%%%%%%%%%%%%%%%%%%%%%%%%%%%%%%%%%%%%%%%%%%%%%%%%%%%%%%%

\newpage
\newpage
\input{sections/appendix}

%%%%%%%%%%%%%%%%%%%%%%%%%%%%%%%%%%%%%%%%%%%%%%%%%%%%%%%%%%%%%%%%%%%%%%%%%%%%%%%
%%%%%%%%%%%%%%%%%%%%%%%%%%%%%%%%%%%%%%%%%%%%%%%%%%%%%%%%%%%%%%%%%%%%%%%%%%%%%%%

\end{document}

%% file: sections/00-abstract.tex
\begin{abstract}
Coreset Selection (CS) aims to identify a subset of the training dataset that achieves model performance comparable to using the entire dataset. 
Many state-of-the-art CS methods select coresets using scores whose computation requires training the downstream model on the entire dataset first and recording changes in the model's behavior on samples as it trains (training dynamics). 
These scores are inefficient to compute and hard to interpret, as they do not indicate whether a sample is difficult to learn in general or only for a specific downstream model. 
Our work addresses these challenges by proposing a score that computes a sample's difficulty using human-understandable textual attributes (concepts) independent of any downstream model. 
Specifically, we measure the alignment between a sample's visual features and concept bottlenecks, derived via large language models, by training a linear concept bottleneck layer and computing the sample's difficulty score using it.
We then use stratified sampling based on this score to generate a coreset of the dataset.
Crucially, our score is efficiently computable without training the downstream model on the full dataset even once, leads to high-performing coresets for various downstream models, and is computable even for an unlabeled dataset.
Through experiments on CIFAR-10/100, and ImageNet-1K, we show that our coresets outperform random subsets, even at high pruning rates, and achieve model performance comparable to or better than coresets found by training dynamics-based methods.
\end{abstract}

%% file: sections/01-introduction.tex
% \vspace{-0.3cm}
\section{Introduction}
\label{sec:introduction}
% Machine learning (ML) pipelines are becoming increasingly intensive regarding their data and compute requirements
Machine learning (ML) pipelines are increasingly demanding more data and compute~\cite{touvron2023llama, achiam2023gpt} to achieve improved performance on various tasks. 
While in line with empirical neural scaling laws~\cite{kaplan2020scaling, hestness2017deep, henighan2020scaling, rosenfeld2019constructive} where a model's performance improves with increasing model and training data size, these improvements come at an unsustainable cost of compute/energy. 
However, recently~\cite{sorscher2022beyond, li2024datacomp} demonstrated that data pruning plays a crucial role in enabling an exponential reduction in test error with increasing dataset size, underscoring the importance of data quality over quantity.

% \begin{wrapfigure}{r}{0.4\textwidth}
%   \centering
%   \includegraphics[width=0.38\textwidth]{imgs/training_dynamics_overview.pdf}
%   \caption{An example image. \label{fig:example}}
% \end{wrapfigure}

\begin{wrapfigure}[9]{r}{0.45\textwidth}
 \vspace{-0.5cm}
  \centering{
  \includegraphics[width=0.43\textwidth]{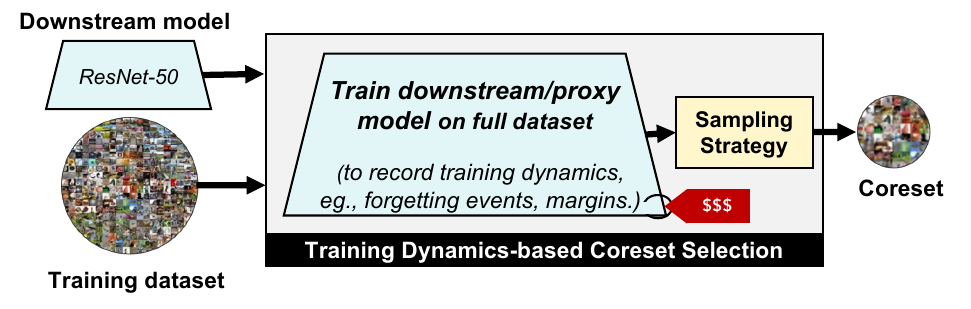}
  }
  \vspace{-0.3cm}
  \caption{\small{Computation of training dynamics-based scores require training the original or a big proxy model on the full dataset, to identify the coreset, making them computationally expensive.} \label{fig:overview_td}
  }
\end{wrapfigure}
Coreset Selection~(CS)~\cite{mirzasoleiman2020coresets, guo2022deepcore,paul2021deep,xia2022moderate,maharana2023d2,zheng2022coverage,choi2024bws} is a data pruning technique that improves the efficiency of model training by pruning a large dataset and retaining only a small subset of representative samples.
Most CS methods work by first using a score to estimate the difficulty/importance of every training sample and then using a sampling strategy that forms the coreset. 
% This is followed by training a downstream model on the coreset leads to similar performance as training that model on the entire dataset. 
%when used for training a model of interest, produces the same performance as when this model is trained on the entire dataset. 
Many state-of-the-art (SOTA) CS methods use the downstream model's training dynamics --- changes in the
model's behavior on a sample over epochs during training, to generate an importance score for each sample. 
While this enables a good estimation of the data importance, it requires training the downstream model on the entire dataset at least once, which can be costly when training a large model on a large dataset
(see Fig.~\ref{fig:overview_td}).
% prohibitive 
% since training a large model on a large dataset even once is computationally challenging.
While~\cite{coleman2019selection} showed that a coreset selected using training dynamics of a mid-sized proxy model (eg, ResNet-18) is effective for a larger downstream model (eg, ResNet-50), training even such a model may not be feasible for large datasets. 
% While \cite{coleman2019selection,zheng2022coverage} showed that a mid-sized proxy model (eg., ResNet-18) could be used to approximate the training dynamics of the actual model of interest (eg., ResNet-50).
% While a mid-sized proxy model may be easier to train than the large model of interest, training even such a model once is still prohibitive when using large datasets. 
% While a mid-a proxy model may be easier to train than the large model of interest, training such a model on a large dataset is still computationally costly. 
% Moreover, the difficulty score computed from these methods is tied to the actual model and may not to estimate the importance of each training sample. 
% However, obtaining information about the training dynamics requires training the model of interest on the full dataset at least once. 
Moreover, since these scores are dependent on training dynamics of a particular downstream model, they are hard to interpret as they do not inform us about the sample's importance for another downstream model (without training it first).
Thus, in our work we tackle the following question:
\emph{``How to efficiently estimate the importance of training samples for CS, in a way that is independent of the downstream model and avoids training that model on the full dataset?}
% How to make such scores more interpretable and aligned to human intuition. 
% training the model of interest on the dataset even once? How can such scores be designed to reflect a sample's difficulty independently of the model?
% "}

% \AM{such scores are interpretable in terms of the model of interest only. what if we don't have the model of interest?}
% \begin{figure}[t]
%   \centering{\includegraphics[width=\columnwidth]{imgs/concept-extraction.png}
%   \caption{{\bf: Automatically generating concept annotation using a language model (LM):} We show the prompt sent to a LM and the concepts it generated for the \textit{golden retriever} class of Imagenet.
%   \AM{Replace the video LLAVA pic with a generic language model.}}
%   }
%   \label{fig:concept-extraction-1}
% \end{figure}

% \begin{figure*}[t]
%   \centering{\includegraphics[width=0.90\textwidth]{imgs/Approach.pdf}
%   \caption{{\bf: Overview of our approach:} We start by learning a linear model that uses concept similarity scores, computed as the alignment between the visual information of a sample and the $k$ concepts for each class (obtained via a language model), to produce a prediction for a training sample. 
%   Based on this, the difficulty score for a training sample is computed as its average margin (i.e., the difference between the likelihood of the correct and the remaining classes) over $T$ training epochs.}
%   }
%   \label{fig:overview}
% \end{figure*}

\begin{figure*}[t]
  \centering{\includegraphics[width=0.9\textwidth]{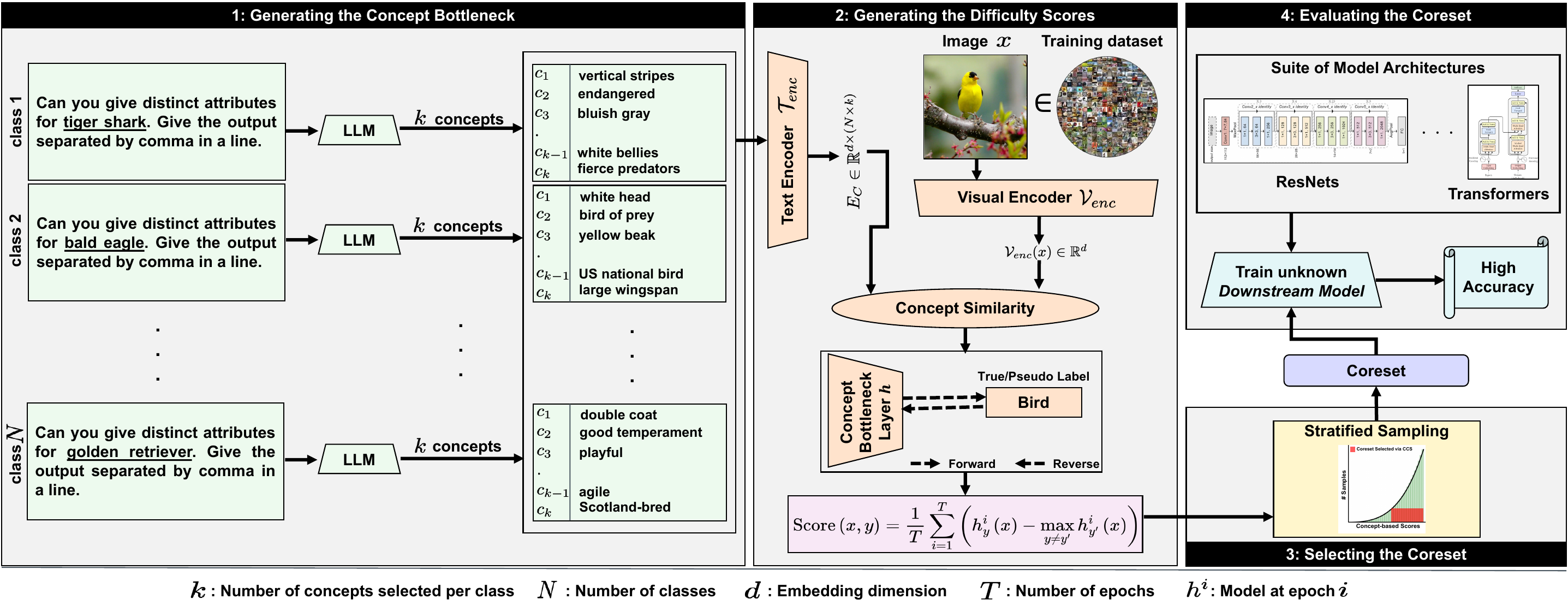}
  \vspace{-0.15cm}
  \caption{\small{\bf Overview of our approach:} 
  We start by prompting an LLM to generate concept annotation for $N$ class label names in the dataset and select $k$ most discriminative attributes (per class) to form the concept bottleneck, which are passed through a text encoder ($\mathcal{T}_{enc}$) to obtain the bottleneck embedding matrix ($E_{C}$). 
  The visual information of a training sample $x$ extracted via the visual encoder ($\mathcal{V}_{enc}$) is then aligned to $E_{C}$ using a linear concept bottleneck layer ($h$) trained for $T$ epochs. 
  Our difficulty score for a sample $x$ is then computed as the average margin (i.e., the difference between the likelihoods of the correct and other classes) over $T$ epochs. 
  Finally, the coreset is selected via stratified sampling and is used to train downstream models. 
  %\AM{update stratified}
  % \AM{Add coreset and multiple models of interest}
  % \emph{Step 1:}   
  %  Automatically generate a concept annotation for all the categories in the dataset using a language model (LM). These concepts are then used as a bottleneck in the CBM. \emph{Step 2:} Learn a linear layer using the concept similarity scores, computed as the alignment between the visual information of a sample and the $k$ concepts for each category, to produce a prediction for a training sample. 
  %  Using this, our concept-based difficulty score for a training sample is computed as the average margin (i.e., the difference between the likelihood of the correct and the remaining classes) over $T$ training epochs. \AM{make it smaller}
   %\AM{Typo in score}
  % \AM{Can we put more discriminative concepts in the picture. They need not be what we are getting from the model.}
  \label{fig:overview}
  }
  % \vspace{-0.15cm}
  }
  
\end{figure*}

To address this, we use Concept Bottleneck Models (CBMs)~\cite{koh2020concept,yuksekgonul2022post}, %which are effective at making ML models more interpretable. 
%and providing a handle for intervening with their predictions. 
which work by mapping a model's input onto a set of human-understandable concepts, referred to as the ``bottleneck'' and use them to make a prediction. 
However, CBMs require concept annotation for every sample in the dataset, which can be costly to obtain. 
Recently~\cite{yang2023language,yan2023learning} showed that off-the-shelf Large Language Models (LLMs) and Vision Language Models (VLMs) can be prompted to obtain concept annotations for training samples without requiring any task-specific fine-tuning (see Fig.~\ref{fig:overview} (block 1)). % shows our prompts to generate the concept bottleneck using an LLM.
%to obtain concept bottleneck for all the class labels 
% categories 
%in the dataset. 
Once the bottleneck is formed, we use a VLM (e.g., CLIP \cite{radford2021learning}) to measure the alignment between the visual features and the concept bottleneck (denoted as concept similarity in Fig.~\ref{fig:overview} (block 2)). 
A linear concept bottleneck layer is then trained to align the visual and concept features while classifying %to make predictions on the 
%training 
samples. 
We then use the average margin of a sample while training the bottleneck layer as our concept-based importance score. % for each training sample. % training dynamics of this bottleneck layer is then utilized to identify the difficulty of each training sample (Fig.~\ref{fig:overview} (right)).
% alignment score is then used to train a linear layer which makes a prediction for the training sample. 
% While training this linear layer we keep track of how a sample's margin changes and use this concept-based score to measure a sample's difficulty (Fig.~\ref{fig:overview} (right)).
Finally, we form the coreset using stratified sampling \cite{zheng2022coverage} (block 3) based on our score, which can then be used for training downstream models (Fig.~\ref{fig:overview} (block 4)). 
% We then use this score along a stratified sampling strategy to 
% Fig.~\ref{fig:overview} provides an overview of our approach. 
% Our method speeds up the computation of the coreset by $15$ times compared to approaches based on the downstream model's training dynamics across three benchmark datasets. %Training the concept bottleneck layer is significantly cheaper than training the downstream model to compute the difficulty score of the samples.
% Moreover, compared to coresets found by other training dynamics-free approaches our coresets lead to downstream models with significantly better accuracy (see Sec.~\ref{sec:experiments}). % shows that our method achieves significantly better results showing its effectiveness for CS.
% to find the coreset s $15$ times faster than training a ResNet-34 model on the Imagenet dataset).
% Thus, unlike previous CS methods, our score which is computable without the knowledge of the architecture and avoids training the downstream model on the entire dataset (even once), makes CS efficient and model agnostic.

We empirically evaluate the effectiveness of our score on three benchmark datasets: CIFAR-10/100 \cite{krizhevsky2009learning} and Imagenet-1K \cite{deng2009imagenet}.
Our results show that downstream models trained on our coresets consistently achieve better accuracy than randomly sampled subsets, especially at high data pruning rates, and achieve performance close to SOTA CS methods. % dependent on the training dynamics of models of interest. 
We also show that our approach is effective for the label-free CS problem where the dataset is unlabeled and leads to models with superior performance compared to SOTA label-free CS methods, especially on Imagenet.
% Our results show that models trained on our coresets exceed performance than random subsets and can even outperform SOTA label-free CS on Imagenet.
Since our CS method is independent of the downstream model, we show that our coreset leads to high performance regardless of the architecture of the downstream model.
% unlike other methods, which require training the downstream model on the dataset first to identify the coresets for a new downstream model. 
Moreover, our method speeds up the computation of the coreset by $\approx 8$ times compared to approaches based on the training dynamics of the downstream model.
%suggesting that our coresets capture representative samples from the dataset with ease. 
% also show that our coresets are transferable to many model architectures showing that our score helps identify difficult samples regardless of the model information. 
%M
% methods based on training dynamics of the models of interest.
Lastly, we show that our concept-based difficulty score provides an intuitive explanation of why examples are easy/hard, independent of the downstream model. 
Our main contributions are summarized below:
\vspace{-0.15cm}
\begin{itemize}[leftmargin=0.4cm]
    \item We propose a concept-based score  that efficiently computes a training data sample's importance for CS without training a downstream model on the full dataset, even once. %, speeding up CS by $\approx$15x compared to SOTA methods.
    % and speeding up CS by approximately 15x compared to SOTA methods.
    
    %We propose a concept-based score for CS for gauging a sample's difficulty in a model agnostic way eliminating the need to train a downstream model on the entire dataset, speeding up CS by $\approx$15 times compared to SOTA methods. 
    
    % We propose a concept-based score for interpretable and efficient CS that eliminates the need for training the downstream model on the entire dataset even once.

    % downstream model agnostic, efficient + 15% speedup
    % 5% on average
    % interpreatbility, mislabeled data detection
    % data centric and emphasize that concepts are also a major contributionn for CS. Mention VLMs will get better at concept extraction helping CS
    
    % We propose a concept-based and model-agnostic data difficulty score for improving the interpretability and efficiency of CS that avoids training a large model on the entire dataset even once.

    % We propose an interpretable and efficiently computable data difficulty score for CS using concept bottlenecks generated by pre-trained LMs that avoids training a large model on the entire dataset even once.

    \item Our coresets improve accuracy by $\approx 5$\% over random subsets at high pruning rates, are competitive to SOTA methods, transfer to various architectures, and are computable without labels.
    
    %Coresets found by our approach are transferable to various model architectures 
    % Our method finds coresets leading to high performance on various downstream model architectures and can do so without even requiring dataset's labels.

    \item We show that using CBMs with LLM-generated concepts makes our score interpretable, enabling a data-centric solution for identifying coresets. 
    %is aligned with human perceived difficulty of samples  
    
    %Our coresets improve accuracy by $5$\% on average over random subsets at high pruning rates, match or outperform SOTA CS methods, particularly on Imagenet at lower pruning rates, and achieve a $15$x speed-up for CS compared to training dynamics-based approaches.
    % We show that our coresets lead to a performance improvement of 5\% on average over random subsets at high pruning rates, is at par or better than SOTA CS methods especially on Imagenet for lower pruning rates, and gives a 15x speed-up over CS by training dynamics based approaches. 
    % achieve better performance than random subsets on three benchmark datasets and achieve performance similar to coresets found by SOTA training dynamics-based CS methods.

\end{itemize}

%% file: sections/02-relatedwork.tex
\section{Related Work}
\label{sec:related-work}
\vspace{-0.15cm}
{\bf Coreset selection (CS).} 
CS improves the efficiency of model training by selecting a subset of influential samples. % and discarding less influential ones.
Various approaches have been proposed to generate such a subset \cite{guo2022deepcore}. 
A popular approach uses influence functions \cite{koh2017understanding,chatterjee1986influential, liu2021influence,schioppa2022scaling} which measures the influence of a sample by considering the effect of removing it from the model's training.
While effective, these approaches are computationally costly due to their dependence on higher-order derivatives. 
Approaches that use the dataset's geometric properties such as are another popular choice for CS \cite{sener2017active,sorscher2022beyond,feldman2020turning,feldman2011unified, huang2019coresets}. 
However, the high computational complexity due to their dependence on pairwise distances between the samples prohibits their use on large datasets. 
Another set of approaches select a subset by either matching the gradients to those computed on the entire dataset \cite{mirzasoleiman2020coresets, killamsetty2021grad} or use training dynamics of a model \cite{toneva2018empirical, pleiss2020identifying,lewis1994heterogeneous,culotta2005reducing,paul2021deep} to compute the importance of a sample. 
However, such approaches require repeated training of the downstream model to produce accurate importance scores.
In comparison, our approach avoids using any knowledge of the downstream model for computing the difficulty scores.
Unlike our work which focuses on data pruning, a different line of work \cite{killamsetty2023milo,tukan2023provable} improves the training efficiency by adaptively selecting the best subset of training data (see App.~\ref{App:additional_rw}).
% efficiently and in an interpretable way.

{\bf Concept-based interpretability approaches.}
Concepts are defined as high-level semantics that refers to the abstract and human-interpretable meanings of the visual data, such as objects, actions, and scenes, as opposed to low-level features like edges or textures~\cite{wu2016valueexplicithighlevel}. 
% These high-level semantic features help interpret visual data in a manner consistent with human understanding. 
Concepts have been used in interpretable computer vision to bridge the gap between human understanding and machine perception in various tasks such as image classification. 
Such interpretability methods can be broadly classified as \textit{post-hoc methods} (do not impose any model constraints) or \textit{by-design} methods (see App.~\ref{App:additional_rw}). 
Concept Bottleneck Models (CBMs) extend interpretable-by-design approaches by using human-understandable attributes as an intermediate layer for predictions, as used in few-shot learning~\cite{lampert2013attribute} and attribute learning~\cite{xu2020attribute,russakovsky2012attribute}.
While interpretable, CBMs reliance on costly annotations and lower accuracy compared to end-to-end models limit their usage.
Post-hoc Concept Bottleneck Models (PCBMs) address these issues by incorporating static knowledge bases (e.g., ConceptNet~\cite{speer2017conceptnet}) and residual connections to boost accuracy~\cite{yuksekgonul2022post}.
Recently~\cite{yang2023language,yan2023learning} incorporated LLMs to identify the concept bottleneck making classification more explainable. 
We build on this and use CBMs for CS. 
% unlike our work which uses them for CS.
% However, unlike our work, these works focus on making classification more explainable rather than which use the concepts for the CS problem, their task was to make classification more explainable. 
% For additional related work.
% See App.~\ref{App:additional_rw} for other related work.

%% file: sections/03-preliminaries.tex
\vspace{-0.1cm}
\section{Preliminaries}
\label{sec:preliminaries}
% Here we formulate the CS problem, discuss some metrics for it and provide an overview of CBMs. % for the it, followed by an overview of the concept bottleneck models.

\vspace{-0.05cm}
\subsection{Coreset selection (CS) problem formulation}
\label{sec:preliminaries_cs}
Consider a classification task and data distribution $P$. 
Let $\mathcal{D} = \{(x_i, y_i)\}_{i=1}^{n}$ denote the dataset of $n$ training examples sampled i.i.d. from the distribution $P$ where $x_i$ denotes the data and $y_i \in \mathcal{Y}$ denotes the label from a set of $N$ classes.
CS \cite{coleman2019selection,zheng2022coverage} aims to find a subset $\mathcal{S}$ of $\mathcal{D}$ consisting of $m \leq n$ samples such that the models trained on $\mathcal{S}$ achieve performance comparable to models trained on $\mathcal{D}$. 
% Formally, the CS problem is %defined as: % follows,
Let $\theta_{\mathcal{D}}$ and $\theta_{\mathcal{S}}$ denote the ``\emph{downstream model}"  trained on $\mathcal{D}$ and $\mathcal{S}$ (coreset), respectively and $\ell$ be the loss function then the CS problem is as follows
\begin{equation}
\label{eq:coreset_selection}
\min_{\mathcal{S}:\mathcal{S} \subset \mathcal{D},|\mathcal{S}|=m} \mathbb{E}_{(x,y) \sim P}[\ell(x,y|\theta_{\mathcal{S}})] - \mathbb{E}_{(x,y) \sim P}[\ell(x,y|\theta_{\mathcal{D}})].
\end{equation}
% where $\theta_{\mathcal{D}}$ and $\theta_{\mathcal{S}}$ denote the ``\emph{downstream model}"  trained on $\mathcal{D}$ and $\mathcal{S}$ (coreset), respectively and $\ell$ is the loss function.

\vspace{-0.05cm}
To find this subset $\mathcal{S}$, previous works have proposed scores that gauge a sample's difficulty for a model. 
%, which are later used to form the coreset. 
Approaches such as max entropy uncertainty sampling \cite{lewis1994heterogeneous, Settles_2012}, and least confidence \cite{culotta2005reducing} estimate difficulty using the uncertainty of the model's predictions on a sample. %, to gauge its difficulty.
Another set of approaches such as $k$-center greedy \cite{sener2017active} uses geometric information of the data to filter out redundant samples. 
Yet, another set of approaches uses training dynamics of the downstream model to estimate the difficulty score. 
Scores such as the forgetting events \cite{toneva2018empirical} computed as the number of times a sample gets misclassified after being correctly classified earlier during model training, and the area under the margin (AUM) \cite{pleiss2020identifying} which identifies mislabeled/difficult samples, fall in this category.
% Based on these score, a sampling strategy is then used to identify $\mathcal{S}$.
% While approaches using training dynamics of the downstream model have achieved SOTA results, the requirement of the knowledge/training of the downstream model or a relatively big proxy model on the entire dataset at least once makes them inefficient, especially for large datasets/models.%, even if done once. 
% This motivates the need of data centric approaches for sample's difficulty independent of the downstream model.
% \AM{Sampling strategy}

\subsection{Concept bottleneck models (CBMs)}
\label{sec:preliminaries_cbm}
Recent advances in language model-guided CBMs utilize an off-the-shelf LLM to obtain concept bottlenecks which are then used to predict the labels. 
These works rely on pre-trained multi-modal models (such as CLIP \cite{radford2021learning}) which consists of a visual encoder $\mathcal{V}_{enc}$ and a text encoder $\mathcal{T}_{enc}$ that can map images and text to a $d$-dimensional representation space. % with dimension $d$. 
Let $C = \{c_1, c_2, \cdots, c_{N_C}\}$ be the set of $N_C$ concepts (bottleneck) generated via an LLM, we can then construct a bottleneck embedding matrix $E_C \in \mathcal{R}^{N_C \times d}$ such that each row of the matrix is mapping of the concept $c \in C$ after passing it through textual encoder $\mathcal{T}_{enc}$.
Based on this, a CBM  \cite{yang2023language} produces a prediction $h(x) = f(g(\mathcal{V}_{enc}(x); E_C))$ for a sample $x$ where $g:\mathbb{R}^{d} \rightarrow \mathbb{R}^{N_C}$ computes the similarity of the visual features to each concept in the bottleneck and $f:\mathbb{R}^{N_C} \rightarrow \Delta$ outputs the probability of each class in the label set $\mathcal{Y}$, where $\Delta$ is a $N$ simplex.
We discuss details of $f$ and $g$ in Sec.~\ref{sec:approach}.

%% file: sections/04-approach.tex
\vspace{-2mm}
\section{Methodology}
\label{sec:approach}
% Here we discuss the key components for computing our concept-based score for CS 
%(which does not rely on knowing/training a model of interest on the large dataset) 
% followed by discussing how to compute it in absence of dataset's labels.
% \AM{change transpose notation}

% \begin{figure}[t]
%   \centering{\includegraphics[width=\columnwidth]{imgs/concept-extraction.png}
%   \caption{{\bf: Extracting Concepts using LLaVA:} We use LLaVA to extract appropriate concepts for dataset classes in an efficient manner. In this figure we show that to retrieve concepts for a particular class of imagenet, \textit{golden retriever}, we prompt the pretrained LLaVA model as follows. 
%   \AM{I used this picture early on. We can create a different one showing the concepts for two similar classes}}
%   }
%   \label{fig:concept-extraction}
% \end{figure}

\vspace{-1mm}
\subsection{Generating the concept bottleneck via off-the-shelf LLMs}
\label{sec:concept_extraction}
Since obtaining data with concept annotation is costly, we use LLMs to generate concept annotation for the samples. 
However, generating attributes (word-level concepts) for all the samples in the dataset via LLMs is still costly, hence we generate word-level concepts \emph{only}\ for class label names. 
This approach was recently shown to be effective at generating the concept bottleneck for interpretable image classification ~\cite{yan2023learning,yang2023language}. 
%every sample in a large dataset for attributes is 
% CBMs require data to be annotated for conceptsManually labeling and designing these attribute concepts can be costly, and does not scale to large numbers of classes.
% The first step of our approach is to extract appropriate attribute-level concepts for different categories/classes of the dataset. 
% To ensure this step is efficient, we extract concepts class-wise and not for every image in a dataset, keeping it consistent with prior work~\cite{yan2023learning,yang2023language}. 
% Furthermore, we preferred to use an open-source LLM over GPT.
In Figure~\ref{fig:overview} (block 1), we present the prompts provided to the LLMs to extract the concepts for various class label names. 
The responses of the LLM are then processed to remove formatting errors to obtain the concept sets. 
%(see App.~\ref{app:concept_extraction} for further details). 
Once the per-class concepts are extracted, we select $k$ discriminative concepts per class (concepts that are unique to a class) to form the concept bottleneck. 
% Since some concepts may appear in more than one class, we only keep concepts unique to a particular class. 
Our final list of $k$ concepts for a class contains its class name and $k-1$ attributes generated by the LLM. 
Details of our prompt design, robustness check of different prompts, and examples of the generated attributes are mentioned in App.~\ref{app:concept_extraction}.
% In Sec.~\ref{sec:ablation}, we show ablation studies using different methods and LLMs for obtaining the concepts, along with the effect of $k$ on CS.  

% While 
% \AM{Mention the computational complexity of extracting via class labels (We note that this step is significantly cheaper than training the model of interest on the entire dataset as required by many previously proposed CS scores.), why LLAVA and not GPT. Mention if there is anything special we did for Imagenet.}
% \AM{Create an appendix to show some class-level concepts and make a picture.}

% \AM{Mention about the use/need of class name here. Discuss some way of extending this to other modalities a little, if possible.}

% \AM{Run the experiment with mean image and 1-shot concepts}

\vspace{-1mm}
\subsection{Concept-based score for CS}
\label{sec:concept_score_based_CS}
We now describe how to use the concept bottleneck to produce a difficulty score for the samples in the dataset. %informing us about each training sample's difficulty in our dataset. 
We start by discussing how we learn the functions $f$ and $g$ described in Sec.~\ref{sec:preliminaries_cbm}
%Then we discuss how to sample a coreset from the dataset based on the proposed score. 
(see Fig.~\ref{fig:overview} (block 2)). % for an overview of the method).
% \AM{we do not use architecture info at all}
We use the dot product between the visual embeddings of an image $x$ i.e. $\mathcal{V}_{enc}(x)$ and the bottleneck embedding matrix $E_C$ to measure the alignment between the visual and textual features %of the concepts 
\cite{yang2023language,yan2023learning}. Concretely, we compute the concept similarity score for a sample $x$ as
% the concept similarity scores for a sample $x$ is computed as 
\begin{equation}
\label{eq:concept_similarity}
g(x; E_C) := \mathcal{V}_{enc}(x) \cdot E_C^\intercal. 
\end{equation}
To map the concept similarity score to a prediction in the label space $\mathcal{Y}$, we propose to use a linear (concept bottleneck layer) predictor denoted by $f$.
Concretely, the function $f$ with parameters $W \in \mathbb{R}^{N \times N_C}$ is given by
\(f(x; W) := g(x; E_C) \cdot W^\intercal\). 
We learn the parameters $W$ using %for the function $f$ using %Using these definitions for $f$ and $g$, we learn $W$ for the function $f$ as
\vspace{-0.1cm}
\begin{equation}
\label{eq:ce_loss}
W^* = \arg \min_{W} \frac{1}{n} \sum_{i=1}^{n} \ell(f(x_i; W), y_i), %\sum_{j=1}^N - y_i^j \log(f(x_i; W)^j).
\end{equation}
% \vspace{-0.1cm}
where $\ell(f(x; W), y) = -\log(f(x; W)_y)$ is the cross-entropy loss.
The output of the concept bottleneck layer is defined as, $h(x):=f(g(x; E_C); W^*).$ % $$= g(\mathcal{V}(x), E_C) \cdot (W^*)^T$.
In practice, we learn $W$ using mini-batch gradient descent by running the optimization for $T$ epochs.
We then compute the difficulty of each training sample using area under the margin (AUM) \cite{pleiss2020identifying} while solving Eq.~\ref{eq:ce_loss}, quantifying the data difficulty as the margin of a training sample averaged over $T$ training epochs.  
Concretely, margin of a sample $(x, y)$ at an epoch $t$ is
\(M^t(x,y) = h^t_y(x) - \max_{y' \neq y} h^t_{y'}(x), \) where $h^t_{y'}(x)$ is the prediction likelihood of the bottleneck layer at epoch $t$ for a class $y'$.
Thus, AUM (concept-based score) is %computed as
\vspace{-0.1cm}
\begin{equation}
\label{eq:aum_true_label}
\mathrm{AUM}(x,y) = \frac{1}{T} \sum_{t=1}^T M^t(x,y).
\end{equation}
% \vspace{-0.1cm}
Recent works \cite{pleiss2020identifying, zheng2022coverage, zheng2024elfs} have demonstrated the effectiveness of AUM for computing a sample's difficulty for CS. % over other approaches including Forgetting. 
However, \cite{zheng2022coverage,zheng2024elfs} compute AUM for a specific downstream model by training it on the entire dataset first, which is computationally costly. 
On the other hand, we integrate AUM with the training of a linear layer $h$, training which is significantly cheaper than training the downstream model (training $h$ for 100 epochs takes only $7$ minutes on Imagenet compared to $8$ hours for training a ResNet-34 for 100 epochs).
Moreover, since our score is independent of the downstream model, our coresets can be used for any downstream model without change, unlike training dynamics-based approaches that require computing their coresets again for different/new downstream models.
% Moreover, as we show in Sec.~\ref{sec:experiments} the coreset obtained through our method (which is independent of the knowledge/training dynamics information of the model of interest $\theta$) is transferable to various model architectures at different pruning rates.

{\bf Sampling training examples to form a coreset.} 
After obtaining data difficulty scores, a crucial step is choosing the samples to form the coreset. 
While many previous works \cite{toneva2018empirical, coleman2019selection} have reported encouraging results keeping the most challenging samples (for our concept-based score this means samples with the smallest margin), recent works \cite{zheng2022coverage,sorscher2022beyond} have shown that this could lead to a catastrophic drop in accuracies after training the downstream model on the coreset, especially when the size of the coreset is small.
This is mainly due to poor sample coverage and potentially mislabeled data in the datasets.
To remedy this, we use Coverage-centric Coreset Selection (CCS) proposed by \cite{zheng2022coverage} (see Alg.~\ref{alg:ccs} in App.~\ref{app:CCS_sampling}) which filters out (potentially) mislabeled samples and uses a stratified sampling approach to form the coreset. % achieving superior result for various coreset sizes.
This technique has been shown to consistently achieve superior results to the random baselines for various coreset sizes. 

\vspace{-0.1cm}
\subsection{Concept-based score for label-free CS}
\label{sec:label_free_CS}
Recently, there has been an interest \cite{zheng2024elfs, maharana2023d2, griffin2024zero} in identifying the representative samples from an unlabeled dataset so as to 1) reduce the samples that need to be labeled/annotated by humans and 2) improve the efficiency of model training by only training the model on a subset of data. 
Our concept-based score can also be effectively utilized for this task with a simple modification. 
Similar to previous works \cite{maharana2023d2,zheng2024elfs,sorscher2022beyond}, we assume that we know the number of classes in the datasets. Additionally, we assume that we also know the names of the classes in the datasets. %(\AM{revisit this} this may be different from dataset class names). 
Previous works have demonstrated that VLMs such as CLIP \cite{radford2021learning} achieve excellent zero-shot performance without requiring fine-tuning on specific datasets. 
We leverage this capability of CLIP models to obtain pseudo-labels for our unlabeled dataset and use them to obtain our difficulty score as follows 
\vspace{-0.1cm}
\begin{equation}
\label{eq:aum_pseudo_label}
\mathrm{AUM}(x, y_{\texttt{pseudo}}) = \frac{1}{T} \sum_{t=1}^T M^t(x,y_{\texttt{pseudo}}),
\end{equation}
% \vspace{-0.1cm}
where for an image $x$ in the dataset $y_{\texttt{pseudo}} = \arg \max_{j \in \mathcal{Y}} \mathcal{V}_{enc}(x) \cdot \mathcal{W}^\intercal_{\texttt{zeroshot}}$ where $\mathcal{W}_{\texttt{zeroshot}} \in \mathbb{R}^{N \times d}$ is a matrix with columns defined as $\mathcal{T}_{enc}(s_j)$ and $s_j = $ ``a photo of a \{\texttt{$j^{th}$ \texttt{class name}}\}" for each class $j \in \mathcal{Y}$ \cite{radford2021learning, wortsman2022robust}.
We use these scores along with CCS \cite{zheng2022coverage} to produce the coreset. 
Similar to \cite{zheng2024elfs, maharana2023d2}, the coreset is then assumed to be annotated by humans and models are trained on it. %this annotated coreset. % and their performance is reported in Sec.~\ref{sec:experiments}. 

%% file: sections/05-experiments.tex
\vspace{-0.1cm}
\section{Experiments}
\label{sec:experiments}
\vspace{-0.1cm}

{\bf Datasets, models, and training:}
We focus on CS for classification tasks on three benchmark datasets namely, CIFAR-10, CIFAR-100 \cite{krizhevsky2009learning}, and Imagenet-1K \cite{deng2009imagenet} datasets consisting of $50000$, $50000$, and $1.28$ million samples spread across $10$, $100$, and $1000$ classes, respectively. 
For CIFAR-10/CIFAR-100, we train a ResNet(RN)-18 as a downstream model and for Imagenet we train ResNet-18/34/50 \cite{he2016deep}, MobileNet \cite{sandler2018mobilenetv2}, DenseNet \cite{huang2017densely}, Wide ResNet \cite{zagoruyko2016wide}, and ViT \cite{dosovitskiy2020image} as downstream models on the coresets.
% For Imagenet, we consider training  for $100$ epochs on the coresets identified by our method. 
We use FFCV~\cite{leclerc2023ffcv} to accelerate training on Imagenet.
We run CS for three trials with different random seeds and report the average of these runs in our tables for various pruning rates where a pruning rate of $90$\% refers to removing $90$\% of the samples. % from the original training dataset.

For generating concepts we use a recently proposed open source model LLaVA~\cite{liu2023llava,liu2023improvedllava}. 
%Extracting the concepts for each class using this model takes 3 seconds per prompt without parallel computation. %In Sec.~\ref{sec:ablation}, we present an ablation of using different VLMs for concept extraction. 
For computing the concept similarity scores between the visual and concept features we used the CLIP \cite{radford2021learning} model (with the ViT B-32 as backbone) following the previous works \cite{yun2022vision,yang2023language,yan2023learning}. % which showed its effectiveness for this task.  
% Specifically, we used the ViT B-32 CLIP model \cite{radford2021learning}. % and present an ablation of using other models in Sec.~\ref{sec:ablation}. 
For computing the pseudo-labels for label-free CS (in Sec.~\ref{sec:label_free_CS}) we used a ViT L-14 CLIP model trained on the DataComp-1B dataset \cite{ilharco_gabriel_2021_5143773}.
%The accuracies of the models trained on the entire training set are $95.44$\% and $78.74$\% for ResNet(RN)-18 on CIFAR-10/100 and $72.4$\% for RN-18, $75$\% for RN-34, and $78.4$\% for RN-50 on Imagenet. 
Further experimental details are mentioned in App.~\ref{app:implementation_details}. % \AM{Fill these}.

% \begin{table}[tb]
%   \caption{Performance of models with different architecture trained on the selected coreset for Imagenet at different pruning rates. We show that results of standard CS (Ours) and label-free CS (Ours-LF) perform better than the models trained on a random subset of data. 
%   }
%   \label{Table:transferability_of_coresets}
%   \centering
%   \small
%   \resizebox{\columnwidth}{!}{
%     \begin{tabular}{c|c|cccc|}
%     \toprule
%     \multirow{2}[1]{*}{Arch.} & \multirow{2}[1]{*}{Method} &  \multicolumn{4}{c|}{\makecell{Pruning Rates}}\\ 
%     & & 30\% & 50\% & 70\% & 90\% \\
%     \midrule
%     \multirow{3}[2]{*}{RN-18} & Random & 71.15 & 68.48 & 63.15 & 44.96 \\
%     \cmidrule(lr){2-6}
%     & Ours & 70.94 & 69.30 & 65.16 & 49.57  \\
%     & Ours (LF) & & & & \\
%     \midrule
%     \multirow{3}[2]{*}{RN-34}&  Random  & 73.37 & 71.71 & 67.85 & 51.29  \\
%     \cmidrule(lr){2-6}
%     & Ours & 73.39 & 72.34 & 69.44 & 55.92  \\
%     & Ours-LF & & & & \\
%     \midrule
%     \multirow{3}[2]{*}{RN-50}&  Random  & 76.06 & 74.44 & 70.50 & 53.56  \\
%     \cmidrule(lr){2-6}
%     & Ours & 76.29 & 75.12 & 72.05 & 58.26  \\
%     & Ours-LF & & & & \\
%     \bottomrule
%     \end{tabular}
%     }
% \end{table}

\input{tables/coreset-performance-tm}
\input{tables/labelfree-coreset-performance-tm}

{\bf CS baselines and methods:}
We compare our method against various baselines and SOTA CS methods. % proposed by previous works. 
1) {\bf Random:} Uniformly select samples from the datasets to form the coreset. Random$_{\mathrm{FFCV}}$ denotes the performance of the models trained on random subsets of Imagenet using FFCV \cite{leclerc2023ffcv}. %the training code based on FFCV \cite{leclerc2023ffcv}.
2) {\bf Entropy \cite{coleman2019selection}:} Selects samples based on entropy computed as the uncertainty of a model's prediction on a sample.
3) {\bf Forgetting \cite{toneva2018empirical}:} Selects samples based on the forgetting score computed as the number of times a sample is misclassified after being correctly classified earlier during training of the downstream model. A higher forgetting score indicates a more challenging sample.
4) {\bf AUM \cite{pleiss2020identifying}:} Selects samples based on their average margin during training of the downstream model i.e., the difference between the target class and the next highest class across the training epochs. Lower AUM indicates a more challenging sample. 
For forgetting, AUM, and our method, we use CCS \cite{zheng2022coverage} to form the coreset whereas for entropy we select samples with the highest entropy as done in \cite{coleman2019selection,zheng2022coverage}.

For label-free CS, we use 
1) {\bf Prototypicality \cite{sorscher2022beyond}:} which first performs k-means clustering in the embedding space of SwAV \cite{caron2020unsupervised} model and ranks samples based on their Euclidean distance to the cluster centers. 
Samples further away from the cluster center are used to form the coreset. 
2) {\bf ELFS \cite{zheng2024elfs}:} estimates the pseudo-labels of the unlabeled samples using a deep clustering approach (using the embedding space of SwAV \cite{caron2020unsupervised} and DINO \cite{caron2021emerging}) and forms the coreset using training dynamics of the downstream model trained on the pseudo-labeled data. 
Crucially, SOTA methods such as forgetting, AUM, and ELFS require training the downstream model on the entire dataset first for CS, unlike our method which is independent of the downstream model. 
While Random and Prototypicality also don't require the downstream model for CS, our results show that our approach surpasses them.
% We report the results for these methods from

\vspace{-0.1cm}
\subsection{Evaluating performance of our concept-based score for CS}
Table~\ref{Table:coreset_performance} shows the accuracy of models trained on coresets found by various approaches on three datasets for the standard CS problem (where the dataset is labeled). 
Our results show that our coresets lead to significantly better performance, even at higher pruning rates, compared to the random subsets.
Moreover, our method which does not have the knowledge of the downstream model or its training dynamics still provides competitive performance to coresets found by the SOTA approaches based on forgetting and AUM, and even outperforms them on Imagenet for smaller pruning rates. 

% Since our approach requires roughly \AM{fill number} minutes compared to \AM{four} hours to generate the coreset via Forgetting Score/AUM on Imagenet, it is able to identify a well performing coreset much more efficiently.
% \begin{figure*}
% \small
% \centering
% \subfigure[easy cifar10]
% {
% \includegraphics[width=0.98\columnwidth]{imgs/cifar10_easy_images.pdf}
% }
% \hfill
% \subfigure[easy cifar100]
% {
% \includegraphics[width=0.98\columnwidth]{imgs/cifar100_easy.pdf}
% }
% \subfigure[challenging cifar10]
% {
% \includegraphics[width=0.98\columnwidth]{imgs/cifar10_difficult_images.pdf}
% }
% \hfill
% \subfigure[challenging cifar100]
% {
% \includegraphics[width=0.98\columnwidth]{imgs/cifar100_difficult.pdf}
% }
% \caption{c}
% \label{fig:easy_difficult_examples}
% \end{figure*}
% \begin{figure*}[t]
%   \centering{\includegraphics[width=\textwidth]{imgs/vizanalysis-tm.pdf}
%   \caption{}
%   \label{fig:easy_difficult_examples}
%   }
% \end{figure*}
% Mention computational advantage to justify efficiency via a table.

% \subsection{Evaluating performance on label-free CS}
Table~\ref{Table:label_free_coreset_performance} shows the accuracy of models trained on coresets when the training set is unlabeled 
(we report the numbers presented by \cite{zheng2024elfs} for previous methods). 
The results show that the random subsets are a competitive baseline for this problem outperforming Prototypicality \cite{sorscher2022beyond}.
Our results also show that our coresets outperform the random subsets for all pruning rates with significant improvements at higher pruning rates.
Compared to ELFS \cite{zheng2024elfs}, our method provides competitive performance and even surpasses it for lower pruning rates on Imagenet, without using any information about the downstream model's architecture or its training dynamics highlighting its effectiveness. % trained on the entire dataset.  
% This highlights the strength of our approach which can be used as an efficient way of estimating the coreset in the label-free scenario, without many modifications. 
% Thus, unlike ELFS \cite{zheng2024elfs} 
% Thus, our approach is effective for standard and label-free CS. 
%for this problem with a simple modification of utilizing the zero-shot predictions from CLIP as pseudo-labels. %This highlights the effectiveness of our approach for CS. % with CCS is both simple and effective for this task. 
% For CIFAR-10/100, pseudo-labels of the training set computed via our approach achieves an accuracy of $98.52$\% and $87.28$\% where as for Imagenet it achieves an accuracy of $79.47$\%
% % 61.7\% \AM{72.89, 79.47} 
% which is better than the best pseudo-label accuracy obtained by the clustering approach in ELFS ($92.5$\% and $66.3$\% on CIFAR-10/100 and $58.8$\% on Imagenet). 
% Moreover, as shown by ELFS, a better stratified sampling approach than CCS (used throughout the paper) can further improve our performance on this problem. 

Next, evaluate the transferability of the coresets found by our approach to downstream models with different architectures including those based on convolutional neural networks, ResNets and ViTs. 
% performance of training downstream models with three different architectures to highlight the effectiveness of our approach for model-agnostic CS.
% first show that our coresets are transferable and lead to accuracy better than training on random subsets for various architectures of the model of interest. 
Results in Table~\ref{Table:transferability_of_coresets} and Table~\ref{Table:transferability_of_coresets_app} (in App.~\ref{app:transferability_of_coresets}) show that our coresets achieve superior performance than random for all the architectures. 
Thus, our concept-based score in conjunction with stratified sampling \cite{zheng2022coverage} which balances the number of easy/hard examples in the coreset is an effective approach for both standard and label-free CS at various pruning rates.

% \begin{table}[tb]
%   \caption{Comparison of the coreset performance on CIFAR-100 with 90\% pruning rate for concept sets generated using different techniques. 
%   }
%   \label{Table:ablation_concept_extraction}
%   \centering
%   \small
%   \resizebox{\columnwidth}{!}{
%     \begin{tabular}{c|c|cccc|}
%     \toprule
%    \makecell{Random} & \makecell{LF?} & \makecell{CW-A} & \makecell{CW-D} & \makecell{1-S}  & \makecell{IL} \\ 
%    \midrule
%     \multirow{2}[1]{*}{44.76} & No &  51.58 &  51.05 & 51.68 & 52.47 \\
%     & Yes & 49.27 & 49.33 & 49.40 & 49.33 \\
%     \bottomrule
%     \end{tabular}
%     }
% \end{table}

\begin{figure*}[t]
  \centering{\includegraphics[width=0.9\textwidth]{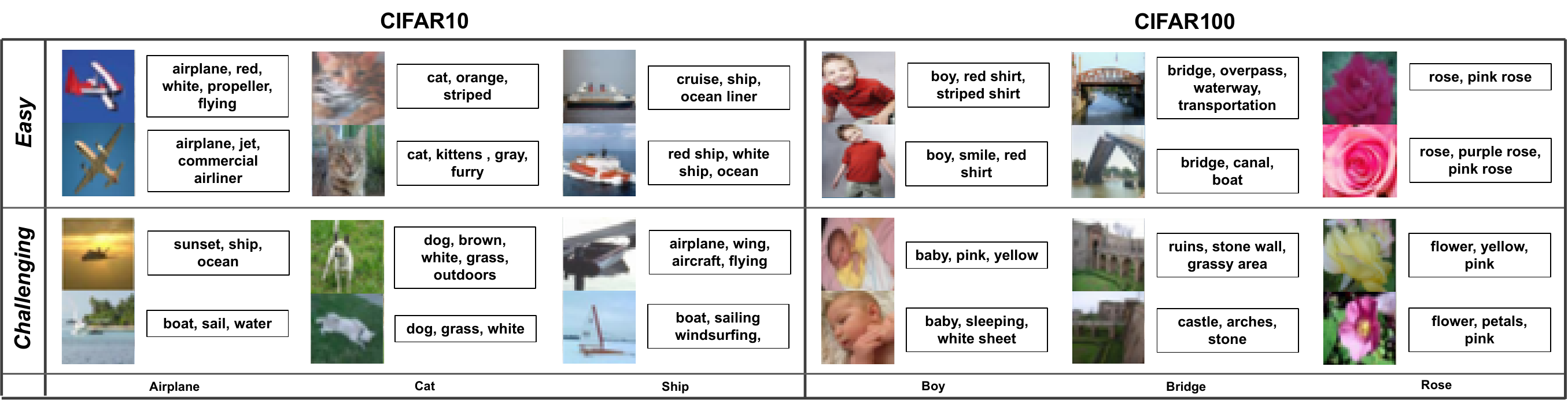}
  \vspace{-0.25cm}
  \caption{\small{Visualizing samples according to our concept-based score for a subset of classes in CIFAR-10/100 showing that easy (challenging) samples are aligned (unaligned) with their assigned label.
  Image-level concepts (in boxes) extracted via LLaVA confirm that easy (challenging) examples are aligned (unaligned) with concepts of their labels, explaining the reason for a high (low) concept-based score.}}
  \label{fig:easy_difficult_examples}
  }\vspace{-0.25cm}
\end{figure*}

\subsection{Evaluating efficiency of CS via our concept-based approach}

% and show 
% This highlights the effectiveness of our concept-based score at producing high performing coresets regardless of the architecture or training dynamics of the model of interest. 

Here we compare the efficiency of our approach at finding coresets compared to approaches based on training dynamics.
%relying on training the downstream model on the entire dataset to obtain the training dynamics-based score. 
For training dynamics-based approaches the time required to find the coreset is dominated by the time needed to train the downstream model on the entire dataset first. 
For example, finding a coreset of Imagenet using a ResNet-34 model takes roughly 8 hours on two A-100 GPUs. % for these approaches.
In comparison, for our approach, extracting concepts via LLMs (block 1 of Fig.~\ref{fig:overview}) takes 3 seconds per class (totaling to 25 minutes for all classes of Imagenet using 2 GPUs), and computing visual/concept features via CLIP and training a linear concept bottleneck layer (block 2 for Fig.~\ref{fig:overview}) takes roughly 30 minutes for Imagenet offering $\approx 8$x speedup for CS over approaches relying on training dynamics of the downstream model.
% Using two A-100 GPUs, our approach can find the coreset in approximately $30$ minutes for the Imagenet dataset giving a $15$x speed up over training dynamics-based approaches. 
% We obtain a similar speed up for CIFAR-10/100 where our method finds the coreset in less than $2$ minutes.
% our approach takes 2hr 12 mins to get concept similarity + 13 mins for LF. First step can be done parallel (16.5 min in 8 jobs). 16.5 +13 ~ 30 mins. CCS takes 8 hrs to train the imagenet model. 8*60/30 = 16
Moreover, since our method is independent of the downstream model architecture we do not need to repeat the CS step for different architectures.
This is in contrast with training dynamics-based methods that require training the downstream model for every  new architecture to find the corresponding best coreset for it.
% \AM{Add time of prompt selection in our approach.}

\input{tables/transferability-coreset-tm.tex}

\subsection{Visualizing easy and challenging samples based on our concept-based score}
\label{sec:visualization}
Here we show the advantage of our concept-based score for assessing the sample's difficulty and how using concepts aids its interpretability. %of our difficulty score. 
We start by visualizing the easiest and the most challenging images (per class) for CIFAR-10/100.
In Fig.~\ref{fig:easy_difficult_examples}, the top row shows the images with the highest concept-based scores (easiest) and the bottom shows the images with the lowest scores (challenging) for a subset of classes in CIFAR-10/100. 
As observed the easiest images are typical images associated with the label where as the challenging images are confusing (and even potentially mislabeled) as they look like images from a different class. % in the same dataset. 
For example, some challenging images in the class ``boy'' from CIFAR-100 are actually images of a baby which is also a class in CIFAR-100. 
Similarly, some challenging images from the class ``cat'' in CIFAR-10 look like images of a dog. 
More examples of such images are presented in Fig.~\ref{fig:easy_difficult_examples_all} in App.~\ref{app:visualization}.
Since the challenging examples seem to be confusing, it shows that our score can identify examples that are ambiguous or mislabeled. 
Such samples may be hard for some ML models to learn and could force them to rely on spurious features, potentially hurting generalization. 
The ability of our score to identify such samples without the downstream model demonstrates its effectiveness for ranking the samples for CS.

% Since the challenging examples are confusing for humans too, {\bf any} ML model will find them hard to learn as well. 
% This shows that our score is well aligned with the human-perceived difficulty of the samples. % and can estimate this difficulty without requiring any information from the training dynamics of the model of interest. 
 
Next, we demonstrate why certain samples get low/high concept-based scores in our approach by extracting concepts specific to these images using LLaVA (note that these concepts are different from the per-class concepts used in the concept bottleneck). 
To generate these, we prompt LLaVA to produce concepts using both the sample image and its class label (see image-level concept extraction in App.~\ref{app:concept_extraction}).
These image-level concepts are shown in the boxes in Fig.~\ref{fig:easy_difficult_examples}.
% of the ass the images along with a text prompt to the LLaVA model and evaluate how well the concepts provided by LLaVA describe the label assigned to the image in the dataset.
As observed in Fig.~\ref{fig:easy_difficult_examples}, image-level concepts provided by LLaVA are related to the class label for easy images whereas they are unrelated for challenging images. 
%This implies that our concept-based score in Eq.~\ref{eq:aum_true_label}, can correctly identify when visual features are aligned (unaligned) with the class labels.
% A high concept-based score as per Eq.~\ref{eq:aum_true_label}, suggests that the visual features of a sample are well aligned with the features of the concepts of the associated class in the CLIP space.
% Since our score in Eq.~\ref{eq:aum_true_label}, is based on similarity between visual features of a sample and those of the  concept sets as per the CLIP model, intuitively, a sample with high score indicates that the LLaVA model describes the image to be similar to the concepts of the sample's category. 
% We indeed find this to be the case as shown in Fig.~\ref{fig:easy_difficult_examples} which shows that for easy images the concepts provided by LLaVA are well aligned with the associated category where as for challenging images there is a mismatch. 
For example, attributes provided by LLaVA for the challenging images of ``airplane'' align more with those of ship (both which are classes in CIFAR-10), and concepts provided for challenging images of ``bridges'' align more with those of castles (both of which are classes in CIFAR-100). 
Since our concept-based score in Eq.~\ref{eq:aum_true_label} assigns a small value for these images, we see that it  correctly captures when a sample's visual information is not aligned with the sample's associated label and vice-versa. 
Thus, explaining why certain examples should be included/excluded from the coreset in an interpretable way independent of the downstream model.
% Moreover, our score can aid in filtering of the datasets by identifying mislabeled/confusing samples which can either be removed or flagged for human evaluation.
% Thus, our score which captures the similarity between the visual embeddings of an images and the concept bottleneck embeddings 

% \AM{It is a cheap way to identify mislabeled examples. We can correct it before model training. poisoning example filtering. Sample is challenging for any ML model}
\vspace{-0.2cm}
\subsection{Ablation studies}
\label{sec:ablation}

Here, we present ablation studies to evaluate various component of our approach. Specifically, we evaluate the effect of different 1)  methods and LLMs used for concept generation, 2) number of concepts per class ($k$) in the bottleneck, 3) CLIP backbones for visual/concept similarity, 4) number of iterations ($T$) for training the concept bottleneck layer. 
Here we use CIFAR-100 with a $90\%$ pruning ratio and train a ResNet-18 on the selected coresets.
Lastly, we also evaluated the effectiveness of our method for CS on tasks beyond object recognition using medical and affective computing tasks. % along with an evaluation on a related problem of adaptive subset selection in App.~\ref{app:adaptive_cs}. 
% A random subset produces an accuracy of $44.76_{\pm 1.58}$ in this setting.
% attribute vs description-level concepts, and finally a comparison between per class and per-image concepts.  

% \input{tables/tab456}
{\bf Comparison of different techniques to generate concepts via LLaVA.}
% {\bf Comparison of attribute-level vs. description based concepts} 
% techniques to generate concepts via LLaVA.}
Tables~\ref{Table:ablation_concept_extraction_classwise} and~\ref{Table:ablation_concept_extraction_imagewise} shows how the performance of models trained on our coresets change when different method are used to generate the concept sets (Fig.~\ref{fig:overview} block 1). 
Since our method uses LLaVA \cite{liu2023llava}, which is a VLM, we compare the performance of models trained on the random subsets and coresets obtained using class-wise concepts (only textual information) and concepts extracted using both visual and textual information.

\input{tables/ablation-concept-extraction-classwise}
For concepts generated using only textual information, we consider two alternatives, namely {\bf class-wise attributes} (CW-A) and {\bf class-wise descriptions} (CW-D). 
While CW-A considers concepts formed by a single or a few words, CW-D consists of longer, more descriptive concepts (eg., a descriptive concept for the class butterfly is ``\emph{a beautiful insect with colorful wings}").
For CW-D, we use a subset of $k$ concepts provided by \cite{yang2023language}, generated via the GPT-3 model. 
Our results show that CW-A performs better than CW-D for both the standard/label-free CS problems. 
% However, generating concepts via GPT-3 model is more expensive than using an open source model. 
% However, attribute-level concepts allow for selecting discriminative concepts which lead to better performance compared to random concepts (see Table~\ref{Table:ablation_k}).
Thus, we used CW-A for all our experiments. %\footnotemark[\value{footnote}].
\footnotetext{Cf. A random subset of CIFAR-100 achieves an accuracy of $44.76_{\pm 1.58}$ at 90\% pruning rate.}
% Thus, we Since longer descriptions were not significantly better in our approach, we consider using attribute level concepts everywhere.
%Hence, considering the simplicity and efficiency of extracting attribute-level for every class, we used concepts sets based only on the textual attributes of classes in our paper.  
% \protect\repnote

\input{tables/ablation-concept-extraction-imagewise}
Next, for generating concepts using both visual and textual information, we consider two alternatives.
The first is a {\bf class-wise one-shot image attribute} approach where we first cluster all images of a class in the embedding space of the CLIP's visual encoder and identify the image whose embedding is the closest to the cluster center (for the label-free setting we use the pseudo-labels of the images during clustering),
%\AM{we have not done this. Need to do it before rebuttal}
% ), 
then we prompt LLaVA to generate attributes using this single image and the class name. 
Once generated, we use $k$ discriminative concepts to form the bottleneck.
The second alternative is the {\bf image-wise} attribute approach, where we use \emph{each} image in the training set and prompt LLaVA to generate per image attributes describing the image. 
Once generated, we sort the concepts based on their frequency of occurrence in a class and use the most frequently occurring discriminative concepts to form the bottleneck.
While the image-level concepts lead to the best coresets, it is slow and costly to prompt LLaVA to generate attributes for all the images in a large dataset such as Imagenet. 

\input{tables/ablation-k-tm.tex}
For CIFAR-100, this process took about nine hours to complete (in comparison CW-A can be extracted in 5 minutes without parallel computation) 
%seconds per class)which may be reduced with exploiting parallel computation) 
which is very costly compared to the small performance gains it provides over other approaches. 
% simpler approaches to obtain concepts.
%\AM{We did not test with single image without clustering since it is not possible to do that without labels. Plus the image we used is in the dataset so result should not be different.}
Lastly, while the class-wise one-shot image attribute approach is better than CW-A, the additional step of clustering can be costly for larger datasets such as Imagenet. 
Based on these results, we used CW-A for concept generation using LLaVA.
% Lastly, we compared the performance of our concepts sets ({\bf class-wise attributes (CW-A)}) to concepts sets which contained more descriptive text ({\bf class-wise descriptions (CW-D)}) rather than attributes for every class (eg., a descriptive concept for the class butterfly is ``\emph{a beautiful insect with colorful wings}").
% For this, we used a subset of $k$ concepts provided by \cite{yang2023language}, which have been extracted using a GPT-3 model. 
% Our results show that the two class-wise concepts perform very similarly on the CS problem. 
% Hence, considering the simplicity and efficiency of extracting attributes for every class, we used concepts sets based only on the textual attributes of classes in our paper.  
% We compare the performance to random subsets and the class-level concepts extracted as described in Sec.~\ref{sec:concept_extraction}. 

{\bf Effect of $k$ and the method for selecting $k$ concepts.} In Table~\ref{Table:ablation_k}, we show how the number of concepts extracted per class label, for creating the bottleneck in block 1 of Fig.~\ref{fig:overview} affects the selection of coresets. 
Once the list of class-wise attribute-level concepts is generated by LLaVA, we can select $k$ concepts per class either randomly or choose concepts unique to a class (discriminative). 
Our results show that using even $k=1$ is sufficient to surpass the performance using a random subset\protect\footnotemark[\value{footnote}]. % (accuracy of $44.76_{\pm 1.58}$). 
This performance increases when we keep discriminative concepts in our concept bottleneck, with $k=5$ achieving the best results.
While the size of the concept bottleneck need not be large to find good coresets, it is helpful to take a sample's visual similarity with a set of concepts rather than a single concept per class. % (or naively the sample's label). 
Thus, we used $5$ discriminative concepts per class to form the bottleneck. % in our experiments.
% \AM{add citation of concept set size}

\input{tables/tab3}
{\bf Effect of using different LLMs/VLMs for concept extraction.}
We evaluated three different VLMs for concept extraction in block 1 of Fig.~\ref{fig:overview}. We used GPT-3 and two open source VLMs namely Phi-3-Mini-4K-Instruct with 3.8 billion parameters, and LLaVA with 7 billion parameters. 
In Table~\ref{Table:ablation_llm_concepts}, we find that the performance of our method remains stable regardless of the LLM used, indicating that even smaller LLMs are effective at generating concepts that produce high performing coresets. 
% This hence make this a not-so compute intensive step of our approach.  of the LLM used for obtaining concepts.

{\bf Effect of using different VLMs for measuring visual and concept similarity.}
Here we evaluated different CLIP backbones to compute the similarity between the visual and concept features used in Eq.~\ref{eq:concept_similarity} (block 2 of Fig.~\ref{fig:overview}). 
% We selected the coreset by measuring visual and concept similarity using different CLIP models. 
Our results in Table~\ref{Table:ablation_clip_backbones} show that our concept-based method achieves significantly better performance than Random\protect\footnotemark[\value{footnote}] for all the backbones with  %  (accuracy of $44.76_{\pm 1.58}$) 
ViT B-32 backbone performing the best. 
% similar performance using various backbones, with ViT B-32 performing the best. 
Thus, we used this backbone for all the experiments in the paper. %in our paper.

{\bf Effect of number of epochs $T$.} Here we evaluated the effect of using different number of epochs $T$ used for training the concept bottleneck layer for computing the concept-based score. Our results in Table~\ref{Table:ablation_T} show that $T \geq 50$ is enough to achieve concept-based scores that lead to selection of high performing coresets. We used $T=100$ in our experiments since that achieves the best performance.

{\bf Evaluation on medical and affective computing tasks.} In App.~\ref{app:cs_for_other_tasks}, we present an evaluation of using our approach for CS for biomedical entity recognition (using BloodMNIST \cite{acevedo2020dataset}) and emotion recognition (using a subset of AffectNet \cite{mollahosseini2017affectnet}) task. 
Superior performance of our approach in Tables~\ref{Table:affectnet-results} and~\ref{Table:bloodmnist-results} on these tasks highlights the effectiveness of our approach for CS on diverse tasks. 

% {\bf Evaluation on adaptive subset selection.}
% In App.~\ref{app:adaptive_cs}, we evaluate the performance of our method on adaptive subset selection problem where the goal is to improve efficiency and convergence of model training not by data pruning but by the selection of a good subset to train on, every few epochs. Our results show that, even on for this application, our concept-based score is an effective method. %yields performance comparable to existing approaches.

% \input{tables/ablation-clip-models}
% \input{tables/ablation-T}

%% file: tables/coreset-performance-tm.tex
\begin{table*}[tb]
  \caption{\small{Comparison of the model's (RN-18 for CIFAR10/100 and RN-34 for Imagenet) test accuracy after training on coresets found by various approaches shows that our coresets lead to significantly better performance than Random achieving competitive results compared to the methods using the downstream model's training dynamics, even for high pruning rates. (Best in each category is highlighted).
  }}
  \label{Table:coreset_performance}
  \centering
  \small
  \resizebox{0.95\textwidth}{!}{
    \begin{tabular}{crcccccccccccc}
    \toprule[1.25pt]
   &\multirow{3}[4]{*}{\textbf{Method}} &  \multicolumn{12}{c}{\makecell{\textbf{Datasets and Pruning Rates}}} \\ 
   \cmidrule(lr){3-14}
   &&  \multicolumn{4}{c}{\makecell{\textit{CIFAR-10}}} & \multicolumn{4}{c}{\makecell{\textit{CIFAR-100}}} & \multicolumn{4}{c}{\makecell{\textit{Imagenet}}} \\
    \cmidrule(lr){3-6} \cmidrule(lr){7-10} \cmidrule(lr){11-14}
    & & $30\%$ & $50\%$ & $70\%$ & $90\%$ & $30\%$ & $50\%$ & $70\%$ & $90\%$ & $30\%$ & $50\%$ & $70\%$ & $90\%$ \\
   
    \midrule
    \multirow{3}{*}{\makecell{\textbf{Needs} \\ \textbf{downstream model} \\ \textbf{training/dynamics}}} & Entropy   & $94.44$ & $92.11$ & $85.67$ & $66.52$ & $72.26$ & $63.26$ & $50.49$ & $28.96$ & $72.34$ & $70.76$ & $64.04$ & $39.04$ \\
    & Forgetting & $\mathbf{95.40}$ & $\mathbf{95.04}$ & $92.97$ & $85.70$ & $\mathbf{77.14}$ & $\mathbf{74.45}$ & $\mathbf{68.92}$ & $\mathbf{55.59}$ & $\mathbf{72.60}$ & $\mathbf{70.89}$ & $66.51$ & $52.28$ \\
    & AUM & $95.27$ & $94.93$ & $\mathbf{93.00}$ & $\mathbf{86.08}$ & $76.84$ & $73.77$ & $68.85$ & $55.03$ & $72.29$ & $70.52$ & $\mathbf{67.78}$ & $\mathbf{57.36}$ \\
    \midrule
    \multirow{3}[2]{*}{\makecell{\textbf{Does not need} \\ \textbf{downstream model} \\ \textbf{training/dynamics}}} & Random & $94.33$ & $93.40$ & $90.94$ & $79.08$ & $74.59$ & $71.07$ & $65.30$ & $44.76$ & $72.18$ & $70.34$ & $66.67$ & $52.34$ \\
    
    &Random$_{\mathrm{FFCV}}$ & - & - & - & - & - & - & - &- & \makecell[tl]{$ 73.37$\\ \tiny{\textcolor{denim}{$\pm 0.08$}}} & \makecell[tl]{$ 71.71$\\ \tiny{\textcolor{denim}{$\pm 0.10$}}} & \makecell[tl]{$ 67.85$\\ \tiny{\textcolor{denim}{$\pm 0.04$}}} & \makecell[tl]{$51.29$\\ \tiny{\textcolor{denim}{$\pm 0.20$}}} \\
    \cmidrule(ll){2-14}
% &\textbf{Ours}  & $\mathbf{94.77}$ & $\mathbf{93.44}$ & $\mathbf{91.80}$ & $\mathbf{84.63}$ & $\mathbf{75.98}$ & $\mathbf{72.22}$ & $\mathbf{66.53}$ & $\mathbf{51.85}$ & $\mathbf{73.39}$ & $\mathbf{72.34}$ & $\mathbf{69.44}$ & $\mathbf{55.92}_{\textcolor{blue}{\pm 17.4}}$ \\
&\textbf{Ours}  
& \makecell[tl]{$\mathbf{94.77}$\\ \tiny{\textcolor{denim}{$\pm 0.09$}}} 
& \makecell[tl]{$\mathbf{93.44}$\\ \tiny{\textcolor{denim}{$\pm 0.61$}}} 
& \makecell[tl]{$\mathbf{91.80}$\\ \tiny{\textcolor{denim}{$\pm 0.21$}}} 
& \makecell[tl]{$\mathbf{84.63}$\\ \tiny{\textcolor{denim}{$\pm 0.24$}}} 
& \makecell[tl]{$\mathbf{75.98}$\\ \tiny{\textcolor{denim}{$\pm 0.26$}}} 
& \makecell[tl]{$\mathbf{72.22}$\\ \tiny{\textcolor{denim}{$\pm 0.22$}}} 
& \makecell[tl]{$\mathbf{66.53}$\\ \tiny{\textcolor{denim}{$\pm 0.42$}}} 
& \makecell[tl]{$\mathbf{51.85}$\\ \tiny{\textcolor{denim}{$\pm 0.29$}}} 
& \makecell[tl]{$\mathbf{73.39}$\\ \tiny{\textcolor{denim}{$\pm 0.12$}}} 
& \makecell[tl]{$\mathbf{72.34}$\\ \tiny{\textcolor{denim}{$\pm 0.13$}}} 
& \makecell[tl]{$\mathbf{69.44}$\\ \tiny{\textcolor{denim}{$\pm 0.17$}}} 
& \makecell[tl]{$\mathbf{55.92}$\\ \tiny{\textcolor{denim}{$\pm 0.02$}}}\\
    \bottomrule[1.25pt]
    \end{tabular}
    }
    \vspace{-0.45cm}
\end{table*}

%% file: tables/labelfree-coreset-performance-tm.tex
% \vspace{-0.15cm}
\begin{table*}[tb]
  \caption{\small{Comparison of the model's test accuracy after training on coresets, found in a {\bf label-free} manner, shows that our coresets lead to better performance than Random and Prototypicality. It is also competitive/better than the coresets found by ELFS (which uses training dynamics of the downstream model).
  }}
  \label{Table:label_free_coreset_performance}
  \centering
  \small
  \resizebox{0.95\textwidth}{!}{
    \begin{tabular}{crcccccccccccc}
    \toprule[1.25pt]
   &\multirow{3}[4]{*}{\textbf{Method}} &  \multicolumn{12}{c}{\makecell{\textbf{Datasets and Pruning Rates}}} \\ 
   \cmidrule(lr){3-14}
   &&  \multicolumn{4}{c}{\makecell{\textit{CIFAR-10}}} & \multicolumn{4}{c}{\makecell{\textit{CIFAR-100}}} & \multicolumn{4}{c}{\makecell{\textit{Imagenet}}} \\
    \cmidrule(lr){3-6} \cmidrule(lr){7-10} \cmidrule(lr){11-14}
    & & $30\%$ & $50\%$ & $70\%$ & $90\%$ & $30\%$ & $50\%$ & $70\%$ & $90\%$ & $30\%$ & $50\%$ & $70\%$ & $90\%$ \\
   
    \midrule

    \multirow{2}{*}{\makecell{\textbf{Needs downstream model} \\ \textbf{training/dynamics}}} & ELFS (SwAV)  & $95.00$ & $94.30$ & $91.80$ & $82.50$ & $76.10$ & $72.10$ & $65.50$ & $49.80$ & $73.20$ & $71.40$ & $66.80$ & $53.40$ \\
    & ELFS (DINO) & $\mathbf{95.50}$ & $\mathbf{95.20}$ & $\mathbf{93.20}$ & $\mathbf{87.30}$ & $\mathbf{76.80}$ & $\mathbf{73.60}$ & $\mathbf{68.40}$ & $\mathbf{54.90}$ & $\mathbf{73.50}$ & $\mathbf{71.80}$ & $\mathbf{67.20}$ & $\mathbf{54.90}$ \\
    \midrule
    \multirow{4}[2]{*}{\makecell{\textbf{Does not need} \\ \textbf{downstream model} \\ \textbf{training/dynamics}}} & Prototypicality  & $94.70$ & $92.90$ & $90.10$ & $70.90$ & $74.50$ & $69.80$ & $61.10$ & $32.10$ & $70.90$ & $60.80$ & $54.60$ & $30.60$ \\
    \cmidrule(ll){2-14}
    
    & Random & $94.33$ & $93.40$ & $90.94$ & $79.08$ & $74.59$ & $71.07$ & $65.30$ & $44.76$ & $72.18$ & $70.34$ & $66.67$ & $52.34$ \\
    &Random$_{\mathrm{FFCV}}$ & - & - & - & - & - & - & - &- & \makecell[tl]{$ 73.37$\\ \tiny{\textcolor{denim}{$\pm 0.08$}}} & \makecell[tl]{$ 71.71$\\ \tiny{\textcolor{denim}{$\pm 0.10$}}} & \makecell[tl]{$ 67.85$\\ \tiny{\textcolor{denim}{$\pm 0.04$}}} & \makecell[tl]{$51.29$\\ \tiny{\textcolor{denim}{$\pm 0.20$}}} \\
    \cmidrule(ll){2-14}
    \cmidrule(ll){2-14}
    % L=14 & \textbf{Ours} & $\mathbf{94.76}$ & $\mathbf{93.47}$ & $\mathbf{91.60}$ & $\mathbf{84.18}$ & $\mathbf{74.75}$ & $\mathbf{71.32}$ & $\mathbf{65.52}$ & $\mathbf{49.27}$ & $\mathbf{73.61}$ & $\mathbf{71.99}$ & $\mathbf{68.42}$ & $\mathbf{53.21}$  \\
    % L-14 datacomp
    % & \textbf{Ours-LF} & $\mathbf{94.81}$ & $\mathbf{93.93}$ & $\mathbf{91.75}$ & $\mathbf{84.02}$ & $\mathbf{74.67}$ & $\mathbf{72.07}$ & $\mathbf{65.50}$ & $\mathbf{49.91}$ & $\mathbf{73.61}$ & $\mathbf{71.99}$ & $\mathbf{68.42}$ & $\mathbf{53.21}$  \\ 
    &\textbf{Ours-LF}  
    & \makecell[tl]{$\mathbf{94.81}$\\ \tiny{\textcolor{denim}{$\pm 0.14$}}} 
    & \makecell[tl]{$\mathbf{93.93}$\\ \tiny{\textcolor{denim}{$\pm 0.13$}}} 
    & \makecell[tl]{$\mathbf{91.75}$\\ \tiny{\textcolor{denim}{$\pm 0.34$}}} 
    & \makecell[tl]{$\mathbf{84.02}$\\ \tiny{\textcolor{denim}{$\pm 0.44$}}} 
    & \makecell[tl]{$\mathbf{74.67}$\\ \tiny{\textcolor{denim}{$\pm 0.23$}}} 
    & \makecell[tl]{$\mathbf{72.07}$\\ \tiny{\textcolor{denim}{$\pm 0.58$}}} 
    & \makecell[tl]{$\mathbf{65.50}$\\ \tiny{\textcolor{denim}{$\pm 0.17$}}} 
    & \makecell[tl]{$\mathbf{49.91}$\\ \tiny{\textcolor{denim}{$\pm 0.96$}}} 
    & \makecell[tl]{$\mathbf{73.61}$\\ \tiny{\textcolor{denim}{$\pm 0.08$}}} 
    & \makecell[tl]{$\mathbf{71.99}$\\ \tiny{\textcolor{denim}{$\pm 0.05$}}} 
    & \makecell[tl]{$\mathbf{68.42}$\\ \tiny{\textcolor{denim}{$\pm 0.21$}}} 
    & \makecell[tl]{$\mathbf{53.21}$\\ \tiny{\textcolor{denim}{$\pm 0.06$}}}\\

    \bottomrule[1.25pt]
    \end{tabular}
    }
     \vspace{-0.2cm}
\end{table*}

%% file: tables/transferability-coreset-tm.tex
\begin{wraptable}[18]{r}{9cm}
% \begin{table}[tb]
\vspace{-0.6cm}
  \caption{\small{
  % Performance of models with different architectures on our coresets for Imagenet. 
  Superior performance of downstream models with various architectures trained on our coresets for Imagenet compared to Random for standard (Ours) and label-free (Ours-LF) CS highlight our method's  effectiveness for CS without knowledge of the downstream model.
  % Results of standard (Ours) and label-free (Ours-LF) CS show that models trained on our coresets perform better than random subsets of data regardless of the architecture of the downstream model. 
  % models trained on our coreset perform better than those trained on a random subsets of data. 
   % \multirow{2}{*}{\makecell{\textbf{Needs Training} \\ \textbf{Dynamics}}
   }}
  \label{Table:transferability_of_coresets}
  \centering
  \small
  \resizebox{0.6\columnwidth}{!}{
    \begin{tabular}{crcccc}
    \toprule[1.25pt]
    \multirow{2}[1]{*}{\makecell{\textbf{Model}\\ \textbf{Architecture}}} & \multirow{2}[1]{*}{\textbf{Method}} &  \multicolumn{4}{c}{\makecell{\textbf{Pruning Rates}}}\\ 
    \cmidrule(ll){3-6}
    & & $30\%$ & $50\%$ & $70\%$ & $90\%$ \\
    \midrule
    \multirow{3}[2]{*}{\textbf{RN-18}} & \textbf{Random} & $71.15_{\scriptscriptstyle \pm0.23}$ & $68.48_{\scriptscriptstyle \pm0.10}$ & $63.15_{\scriptscriptstyle \pm0.19}$ & $44.96_{\scriptscriptstyle \pm0.50}$ \\
    \cmidrule(lr){2-6}
    & \textbf{Ours-LF} & $71.21_{\scriptscriptstyle \pm0.09}$ & $68.77_{\scriptscriptstyle \pm0.13}$ & $63.76_{\scriptscriptstyle \pm0.09}$ & $47.50_{\scriptscriptstyle \pm0.10}$\\
    & \textbf{Ours} & $70.94_{\scriptscriptstyle \pm0.19}$ & $69.30_{\scriptscriptstyle \pm0.84}$ & $65.16_{\scriptscriptstyle \pm0.04}$ & $49.57_{\scriptscriptstyle \pm0.16}$  \\
    
    \midrule
    \multirow{3}[2]{*}{\textbf{RN-50}} & \textbf{Random} & $76.06_{\scriptscriptstyle \pm0.11}$ & $74.44_{\scriptscriptstyle \pm0.04}$ & $70.50_{\scriptscriptstyle \pm0.02}$ & $53.56_{\scriptscriptstyle \pm0.13}$ \\
    \cmidrule(lr){2-6}
    & \textbf{Ours-LF} & $76.54_{\scriptscriptstyle \pm0.08}$ & $74.84_{\scriptscriptstyle \pm0.03}$ & $71.10_{\scriptscriptstyle \pm0.09}$ & $55.22_{\scriptscriptstyle \pm0.92}$\\
    & \textbf{Ours} & $76.29_{\scriptscriptstyle \pm0.10}$ & $75.12_{\scriptscriptstyle \pm0.03}$ & $72.05_{\scriptscriptstyle \pm0.09}$ & $58.26_{\scriptscriptstyle \pm0.46}$  \\

    \midrule
    \multirow{3}[2]{*}{\textbf{DenseNet}} & \textbf{Random} & $76.55_{\scriptscriptstyle \pm0.11}$ & $75.31_{\scriptscriptstyle \pm0.15}$ & $71.85_{\scriptscriptstyle \pm0.20}$ & $56.27_{\scriptscriptstyle \pm0.21}$ \\
    \cmidrule(lr){2-6}
    & \textbf{Ours-LF} & $76.94_{\scriptscriptstyle \pm0.23}$ & $75.52_{\scriptscriptstyle \pm0.11}$ & $72.28_{\scriptscriptstyle \pm0.14}$ & $58.16_{\scriptscriptstyle \pm0.04}$\\
    & \textbf{Ours} & $76.80_{\scriptscriptstyle \pm0.22}$ & $75.94_{\scriptscriptstyle \pm0.03}$ & $73.40_{\scriptscriptstyle \pm0.27}$ & $60.93_{\scriptscriptstyle \pm0.19}$  \\

    \midrule
    \multirow{3}[2]{*}{\textbf{Wide Resnet}} & \textbf{Random} & $77.39_{\scriptscriptstyle \pm0.19}$ & $75.32_{\scriptscriptstyle \pm0.12}$ & $71.51_{\scriptscriptstyle \pm0.10}$ & $54.78_{\scriptscriptstyle \pm0.83}$ \\
    \cmidrule(lr){2-6}
    & \textbf{Ours-LF} & $77.83_{\scriptscriptstyle \pm0.08}$ & $75.77_{\scriptscriptstyle \pm0.09}$ & $71.97_{\scriptscriptstyle \pm0.08}$ & $57.34_{\scriptscriptstyle \pm0.58}$\\
    & \textbf{Ours} & $77.48_{\scriptscriptstyle \pm0.05}$ & $76.32_{\scriptscriptstyle \pm0.03}$ & $73.14_{\scriptscriptstyle \pm0.03}$ & $59.75_{\scriptscriptstyle \pm0.70}$  \\
    
    \bottomrule[1.25pt]
    \end{tabular}
    }
     
% \end{table}
\end{wraptable}

%% file: tables/ablation-concept-extraction-classwise.tex
\begin{wraptable}[8]{r}{4.65cm}
% \begin{table}[t]
\vspace{-0.65cm}
  \caption{\small{
  Effect of using concepts obtained using only textual information (attributes vs. descriptions) to form the concept bottleneck layer \protect\footnotemark{}. 
  % \protect\repnote
  % \protect\footnote{Some footnote text}
  %\repnote. 
  % (cf. $44.76_{\scriptscriptstyle \pm1.58}$ for a Random subset.)
  % Coreset performance on CIFAR-100 for $90\%$ pruning rate for concepts generated using different methods. 
  %A random subset of CIFAR-100 achieves an accuracy of $44.76_{\scriptscriptstyle \pm1.58}$ at this rate. (Class-wise attributes approach is used for in rest of the paper.)
  }}
  % }
  \label{Table:ablation_concept_extraction_classwise}
  \centering
  \small
  \resizebox{0.32\columnwidth}{!}{
    \begin{tabular}{rcc}
    \toprule[1.25pt]
   & \makecell{{\textbf{class-wise}} \\ {\textbf{attributes}}} & \makecell{{\textbf{class-wise}} \\ {\textbf{descriptions}}} \\ 
   \midrule
   \makecell{{\textbf{Ours}}} &  $51.85_{\scriptscriptstyle \pm0.51}$ &  $51.05_{\scriptscriptstyle \pm0.71}$ \\
   \makecell{{\textbf{Ours-LF}}} & $49.91_{\scriptscriptstyle \pm0.96}$ & $49.85_{\scriptscriptstyle \pm0.83}$ \\
    \bottomrule[1.25pt]
    \end{tabular}
    }
% \end{table}
\end{wraptable}

%% file: tables/ablation-concept-extraction-imagewise.tex
\begin{wraptable}[7]{r}{5.5cm}
% \begin{table}[t]
\vspace{-0.65cm}
  \caption{\small{
  Effect of using concepts generated by using both visual and textual information to form the concept bottleneck layer \protect\footnotemark[\value{footnote}]. 
  % \protect\repnote
  % (cf. $44.76_{\scriptscriptstyle \pm1.58}$ for a Random subset.)
  % Coreset performance on CIFAR-100 for $90\%$ pruning rate for concepts generated using different methods. 
  %A random subset of CIFAR-100 achieves an accuracy of $44.76_{\scriptscriptstyle \pm1.58}$ at this rate. (Class-wise attributes approach is used for in rest of the paper.)
  }}
  % }
  \label{Table:ablation_concept_extraction_imagewise}
  \centering
  \small
  \resizebox{0.4\columnwidth}{!}{
    \begin{tabular}{rcc}
    \toprule[1.25pt]
    & \makecell{{\textbf{class-wise one-shot}} \\ {\textbf{image attributes}}}  & \makecell{{\textbf{image-wise}} \\ {\textbf{attributes}}} \\ 
    
   \midrule
   \makecell{{\textbf{Ours}}} &  $51.68_{\scriptscriptstyle \pm0.45}$ & $52.47_{\scriptscriptstyle \pm0.39}$ \\
   \makecell{{\textbf{Ours-LF}}}& $50.22_{\scriptscriptstyle \pm0.16}$ & $51.05_{\scriptscriptstyle \pm0.93}$ \\
    \bottomrule[1.25pt]
    \end{tabular}
    }
% \end{table}
\end{wraptable}

%% file: tables/ablation-k-tm.tex
% \vspace{-0.1cm}
\begin{wraptable}[7]{r}{7.cm}
 \vspace{-0.6cm}
  \caption{\small{
  Effect of the number of per-class concepts $k$ and the method of selecting $k$ attribute-level concepts from those generated by LLaVA on the accuracy of models trained on the coreset\protect\footnotemark[\value{footnote}]. %(cf. $44.76_{\scriptscriptstyle \pm1.58}$ for a Random subset.)
  % Effect of the number of concepts $k$ (per class) and the method of selecting $k$ attribute-level concepts from the concepts generated by LLaVA vs. the accuracy of models trained on the coresets of CIFAR-100 at $90\%$ pruning rate.
  % A random subset of CIFAR-100 achieves an accuracy of $44.76_{\scriptscriptstyle\pm 1.58}$ at this rate.
  % \AM{Add results for CIFAR10 and Imagenet.}
  }}
  \label{Table:ablation_k}
  \centering
  \small
  \resizebox{0.5\columnwidth}{!}{
    \begin{tabular}{rccc}
    \toprule[1.25pt]
    \textbf{Concept selection}  & $k$=1 & $k$=5 & $k$=10  \\ 
    \midrule
    % Random Concepts & $48.78_{\pm0.96}$ & $50.15_{\pm1.64}$	& $50.39_{\pm0.72}$ \\
    % Discriminative Concepts & $51.42_{\pm0.18}$	& {\bf $51.58_{\pm0.51}$} & $51.22_{\pm0.72}$ \\
    \textbf{Random concepts} & $48.78_{\scriptscriptstyle \pm0.96}$ & $50.15_{\scriptscriptstyle \pm1.64}$	& $50.39_{\scriptscriptstyle \pm0.72}$ \\
    \textbf{Discriminative} & $51.42_{\scriptscriptstyle \pm0.18}$	& {$\mathbf{51.85}_{\scriptscriptstyle \pm0.29}$} & $51.22_{\scriptscriptstyle \pm0.72}$ \\
    \bottomrule[1.25pt]
    \end{tabular}
    }
% \end{table}
\end{wraptable}

%% file: tables/tab3.tex
\begin{table*}[t]
  \centering

  % --- Table 1: LLMs ---
  \begin{minipage}[t]{0.32\linewidth}
    \centering
    \caption{\small{Effect of using different LLMs for generating concepts used in the concept bottleneck\protect\footnotemark[\value{footnote}].}}
    \label{Table:ablation_llm_concepts}
    \small
    % \resizebox{0.9\linewidth}{!}{
      \begin{tabular}{cc}
        \toprule[1pt]
        \textbf{LLM} & Accuracy \\
        \midrule
        % \Tstrut Random & $44.76_{\scriptscriptstyle\pm 1.58}$ \\
        \Tstrut \textbf{GPT-3} & $51.05_{\scriptscriptstyle \pm0.71}$ \\
        \Tstrut \textbf{Phi-3} & $51.30_{\scriptscriptstyle \pm0.26}$ \\
        \Tstrut \textbf{LLaVA} & $51.85_{\scriptscriptstyle \pm0.29}$ \\
        \bottomrule[1pt]
      \end{tabular}
    % }
  \end{minipage}
  \hfill
  % --- Table 2: CLIP Backbones ---
  \begin{minipage}[t]{0.32\linewidth}
    \centering
    \caption{\small{Effect\protect\footnotemark[\value{footnote}] of using various CLIP backbones for computing the visual-concept similarity in Eq.~\ref{eq:concept_similarity}.}}
    \label{Table:ablation_clip_backbones}
    \small
    % \resizebox{0.9\linewidth}{!}{
      \begin{tabular}{cc}
        \toprule[1pt]
        \textbf{CLIP backbone} & Accuracy \\
        \midrule
        % Random & $44.76_{\scriptscriptstyle\pm 1.58}$ \\
        \textbf{ViT B-16} & $48.89_{\scriptscriptstyle \pm1.01}$ \\
        \textbf{ViT L-14} & $49.54_{\scriptscriptstyle \pm1.82}$ \\
        \textbf{ResNet-50} & $50.56_{\scriptscriptstyle \pm1.67}$ \\
        \textbf{ViT B-32} & $51.85_{\scriptscriptstyle \pm0.29}$ \\
        \bottomrule[1pt]
      \end{tabular}
    % }
  \end{minipage}
  \hfill
  % --- Table 3: Epochs T ---
  \begin{minipage}[t]{0.32\linewidth}
    \centering
    \caption{\small{Effect\protect\footnotemark[\value{footnote}] of using different number training epochs $T$ for computing the score in Eq.~\ref{eq:aum_true_label}.}}
    \label{Table:ablation_T}
    \small
    % \resizebox{0.9\linewidth}{!}{
      \begin{tabular}{cc}
        \toprule[1pt]
        \textbf{$T$} & Accuracy \\
        \midrule
        \textbf{10} & $46.63_{\scriptscriptstyle\pm 0.08}$ \\
        \textbf{20} & $48.40_{\scriptscriptstyle \pm0.10}$ \\
        \textbf{50} & $50.50_{\scriptscriptstyle \pm0.31}$ \\
        \textbf{100} & $51.85_{\scriptscriptstyle \pm0.29}$ \\
        % \textbf{200} & $51.06_{\scriptscriptstyle \pm 0.54}$ \\
        \bottomrule[1pt]
      \end{tabular}
    % }
  \end{minipage}
\vspace{-0.5cm}
\end{table*}

%% file: sections/06-conclusion.tex
\vspace{-0.2cm}
\section{Conclusion}
\label{sec:conclusion}
\vspace{-0.2cm}
CS finds representative samples from a large dataset, training models on which leads to models with accuracy similar to the models trained on the entire dataset. 
In this work, we proposed a scoring mechanism based on concept bottlenecks that allows us to compute the difficulty of a sample in terms of interpretable concepts.
This method is independent of the downstream model and avoids training the downstream model on the full dataset even once. % of interest.
%which will eventually be trained on the coreset. 
Our experiments show that training downstream models on coresets selected using our approach leads to better performance than random subsets and achieves accuracy similar to or better than the SOTA approaches based on training dynamics of the downstream model, for both the standard and label-free CS problem. 
Moreover, our score provides an intuitive explanation of a sample's difficulty at a dataset level, independent of the downstream model. 
%of the downstream model.

{\bf Limitations:}
While we evaluated various LLMs for concept generation and showed their effectiveness for computing concept-based score, it is possible that for some classes LLMs produce very noisy and non-discriminative concepts, leading to poor concept-based scores. 
Thus, developing methods to prompt the LLMs to generate discriminative concepts to be used as the concept bottleneck is a crucial research direction and is left for future works.
Another limitation is the dependence of our score on the VLMs for computing similarity between visual and concept features. Although we evaluated our approach on a diverse set of tasks including a biomedical task and various CLIP-based VLMs, it is possible that these similarity scores may not work well for certain tasks, warranting additional fine-tuning to make the similarity scores meaningful. 
Hence, in practice, additional time/compute may be required to identify or fine-tune a VLM for CS.
%specific task at hand. 
% Moreover, while class-wise concept extraction is efficient, image-level concepts are more informative for a sample's difficulty. 
% Thus, improving the efficiency of concept extraction can help generate better concept bottlenecks. 
% and tuning the prompt for LLMs/VLMs to incorporate the feedback from score computation or CS can help generate better concept bottlenecks leading to a better estimate of a sample's difficulty.
% While these are important research directions we leave them for future work.
% In our work, the concept extraction from LLM is treated as a pre-processing step, independent of the data difficulty score computation and CS. 
% However, 
% before selecting the coreset i focused on extracting concepts from a pre-trained LLM for all classes in our dataset 
% Mention that we could use prompt tuning to incorporate the feedback of coreset performance back into generation of concept bottlenecks. But since LLAVA concepts require post processing it was hard to do it in the current work. 

%% file: sections/appendix.tex
\appendix
\onecolumn
\begin{center}
{\LARGE \bf Appendix}
\end{center}

We present additional related work in Appendix~\ref{App:additional_rw}. Then we describe the details of our methodology for extracting the concepts from LLaVA in Appendix~\ref{app:concept_extraction} and present additional experiments and implementation details of our experiments in Appendix~\ref{app:implementation_details} including the algorithm for stratified sampling used in our work in Appendix~\ref{app:CCS_sampling}.

\if0
\section{TODO (ignore)}
\begin{itemize}
    \item Make concept-extraction few-shot and more efficient. (Check if class-level concepts are good. If not go for a few-shot image level.)
    \begin{itemize}
        \item Top-k concepts extracted with BERT/LLAVA with class names.
        \item Labo concept selection.
    \end{itemize}

    \item Scale the method to Imagenet and other Image datasets. 

    \item Ablation studies
     \begin{itemize}
        \item Number of top concepts k to keep.
        \item Linear probing vs Non-linear model training with MLPs.
        \item Submodular optimization for top-k concept extraction. 
        \item Evaluate coresets on other model architectures.
        \item Evaluate concept quality with GPT, LLAVA, etc. (Use concise + LABO)
        \item Add additional baselines of Kmeans and other methods.
        \item Show results with image-level concepts compared to class-wise concepts.
     \end{itemize}

     \item Label-free approach: Use CLIP zero-shot predictions as pseudo-labels for the label-free approach. (Additionally, Label noise + feature noise in images and show the Robustness of concepts to these noises.

     \item Interpretability of concept/weight scores + Why does our score not match the AUM score exactly?

     \item Find coreset for a target task from a dataset without labels. Cats and Dogs from a general dataset.

     \item Comparison of vit B-32 and L-14?

     \item Run CS via D2 Pruning.

     \item Compare coresets selected by different methods.
     
\end{itemize}
\fi

\section{Additional related work}
\label{App:additional_rw}

{\bf Concept-based interpretability:} 
Interpretability methods can be broadly classified as \textit{post-hoc methods} (do not impose any model constraints) or \textit{by-design} methods. 
Post-hoc methods include Gradient-weighted Class Activation Mapping approaches~\cite{bau2017network,selvaraju2017grad,mu2020compositional,hernandez2021natural} that trace network gradients to identify the input areas that guide predictions and Explanation Generation methods~\cite{singh2023explaining,nishida2022improving,kim2018textual,hendricks2016generating} that require models to produce explanations for visual tasks by conditioning their predictions on captioning models or incorporating visual evidence to ground explanations~\cite{hendricks2018grounding,park2018multimodal}.
Interpretable-by-design methods, such as Prototype methods, optimize a metric space where classifications are based on distances to class prototypes, identifying important input regions but often obscuring their semantic content~\cite{nauta2021neural,chen2019looks,snell2017prototypical,satorras2018few,vinyals2016matching}. 

Concept Bottleneck Models (CBMs) are a part of interpretable-by-design approaches that use human-understandable attributes as an intermediate layer for predictions.
A recent advancement, Computational Derivation Learning (CompDL), utilizes a CBM architecture by applying a linear layer over CLIP scores between expert-designed concepts and images, improving evaluation of how well CLIP grounds concepts~\cite{yun2022vision}.  
Post-hoc Concept Bottleneck Models (PCBMs) were recently proposed to ease the requirement of CBMs to rely on costly concept annotations and improve their accuracy compared to end-to-end models. 
However, PCBMs are limited by the coverage of knowledge bases, making them unsuitable for large-scale or domain-specific tasks or fine-grained classification, and their residual predictors can undermine interpretability by blending CBMs with end-to-end models.

High-level semantic-driven descriptions are also used to guide data augmentation to build an informative set~\cite{wickramanayake2021explanation} to make model training efficient with a good enough training set. 
Prior works use external knowledge bases to obtain these textual semantic concepts to guide vision models~\cite{bujwid2021large,kil2021revisiting,roth2022integrating,shen2022k}. 
Thus, the use of concepts has been shown to improve interpretability in various domains. However, to the best of our knowledge, we are the first ones to propose a concept-based score for the CS problem and show its competitiveness to SOTA model training dynamics-dependent approaches.

{\bf Adaptive subset selection:} These works focus of improving the training convergence and efficiency of model training by selecting a new subset of data from the whole dataset, every epoch, while training a downstream model. 
Thus, unlike our work, these works do not prune the dataset but rather keep selecting small, potentially non-overlapping, subsets every few epochs.
Both \cite{killamsetty2023milo, tukan2023provable}, propose ways to select subsets every epoch or every few epochs in a way that does not use a downstream model unlike some other works such as GradMatch \cite{killamsetty2021grad}, which select subsets based on the downstream model. 
While our work does not target adaptive subset selection, we evaluated how well our approach performs without any change on this problem in App.~\ref{app:adaptive_cs}. Our results show that, even on for this application, our concept-based score is an effective method. %yields performance comparable to existing approaches.

\section{Details for concept set generation}
\label{app:concept_extraction}
\noindent\textbf{Prompt Selection: }To extract concepts for our approach, we only use the class labels in the prompt as can be seen in Figure~\ref{fig:overview}. 
The prompt, ``Can you give distinct attributes for $ \left< \text{class\;name}\right>$. Give the output separated by a comma in the line.'' instructs the VLM not only to provide distinct keywords but also adds formatting instructions. 
However, despite the instructions included in the prompt, LLaVA outputs are not always formatted well, often containing duplicate entries, mismatched commas and braces, and sometimes having a detailed explanation before the keywords. 
To remedy this we run the LLaVA output through a simple post-processing script and use regular expressions to clean the LLaVA outputs. 
For our experiments where we perform ablation of various concept-bottleneck generation methods~(Tables~\ref{Table:ablation_concept_extraction_classwise},~\ref{Table:ablation_concept_extraction_imagewise}), we also use two more concept generation methods, one is one-shot image-based class concepts and the second is image-level concept generation. 
For the former, where we select one representative image per class via clustering, we prompt LLaVA as follows, ``$ \left< \text{image}\right>$ Can you give distinct attributes for such an image of $ \left< \text{class\;name}\right>$. Give the output separated by a comma in the line.''
And, to get concepts for every image of a class, we use a similar prompt as follows, ``$ \left< \text{image}\right>$ Can you give distinct visual attributes for this image of $ \left< \text{class\; name}\right>$. Give the output separated by a comma in the line.'' Each LLaVA prompt request on a single A-100 GPU takes approximately 3 seconds. 

\noindent\textbf{Alternative VLMs for Concept Generation: } We leverage LLaVA as our choice of VLM for concept generation, however in Table~\ref{Table:ablation_llm_concepts}, we also compared against concepts extracted from GPT~\cite{yan2023learning} and another smaller open source VLM (Phi-3-Mini-4K-Instruct). 
We see comparable performance of our CS method with concepts extracted from these models.
% those extracted from LLaVA. 
% Moreover, LLaVA is an open-source model whereas GPT is not. 
We also experimented with retrieving concepts via another recent method, SpLiCE~\cite{bhalla2024interpreting} which uses a linear optimization for sparse concept decomposition.
However, a major limitation of SpLiCE is that similar to image-wise attributes as used in Table~\ref{Table:ablation_concept_extraction_imagewise} it is a costly approach (SpLiCE  can take up to 3 hours for $50,000$ images, significantly slower than generating class-level concepts from LLaVA).  
% \AM{TM, Can you explain more why it is a costly approach and we did not use it?}. 

% Talk about the prompts used (one-shot, image-level)
% Talk about alternatives to concept extraction; e.g. SPLICE, and why we did not select that. 
% 
%
\section{Additional experiments and implementation details}
\label{app:implementation_details}

\subsection{Visualizing easy/challenging samples based on concept-based score}
\label{app:visualization}
\begin{figure*}
\small
\centering
\subfigure[Easy images from CIFAR-10]
{
\includegraphics[width=0.45\columnwidth]{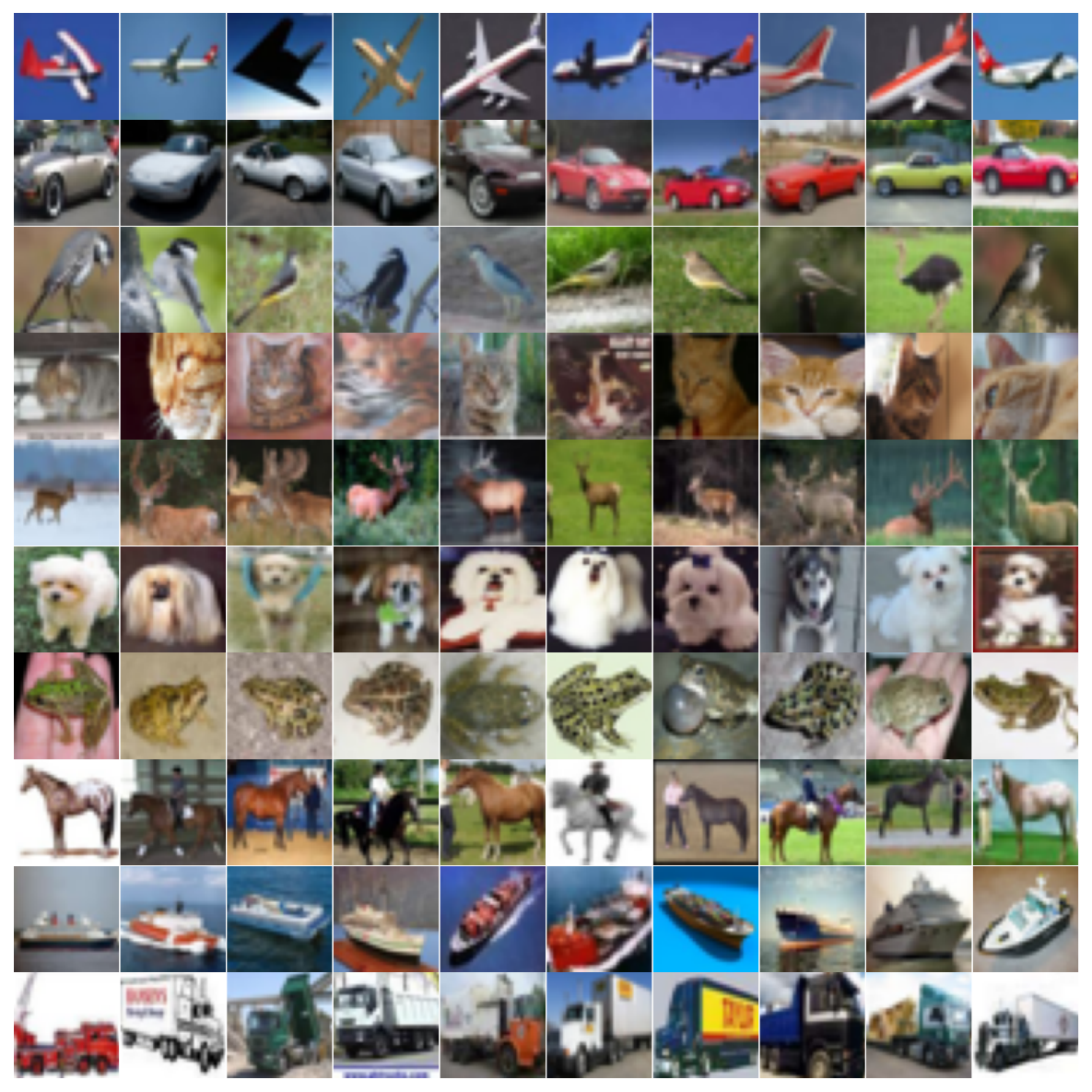}
}
\hfill
\subfigure[Challenging images from CIFAR-10]
{
\includegraphics[width=0.45\columnwidth]{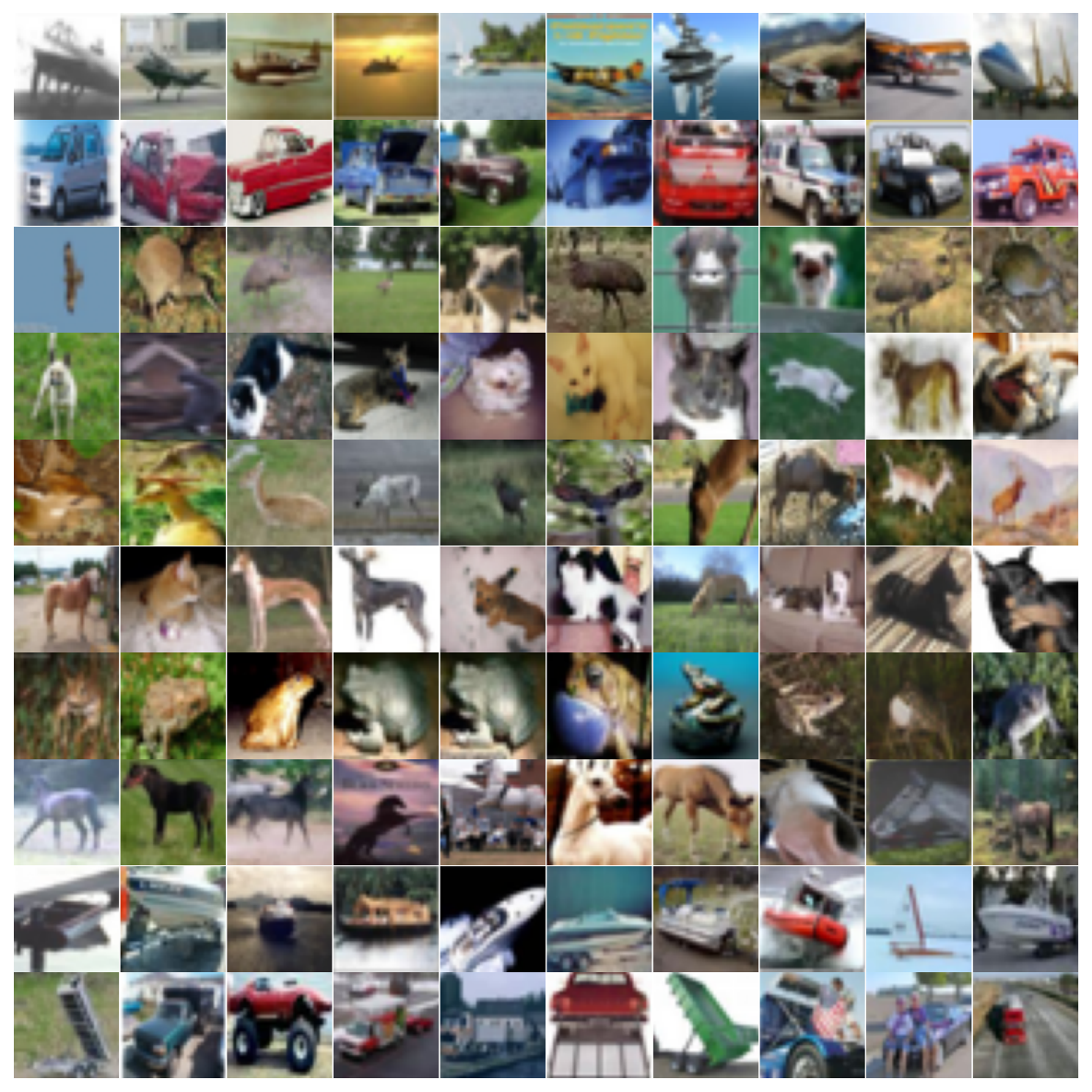}
}
\subfigure[Easy images from CIFAR-100]
{
\includegraphics[width=0.45\columnwidth]{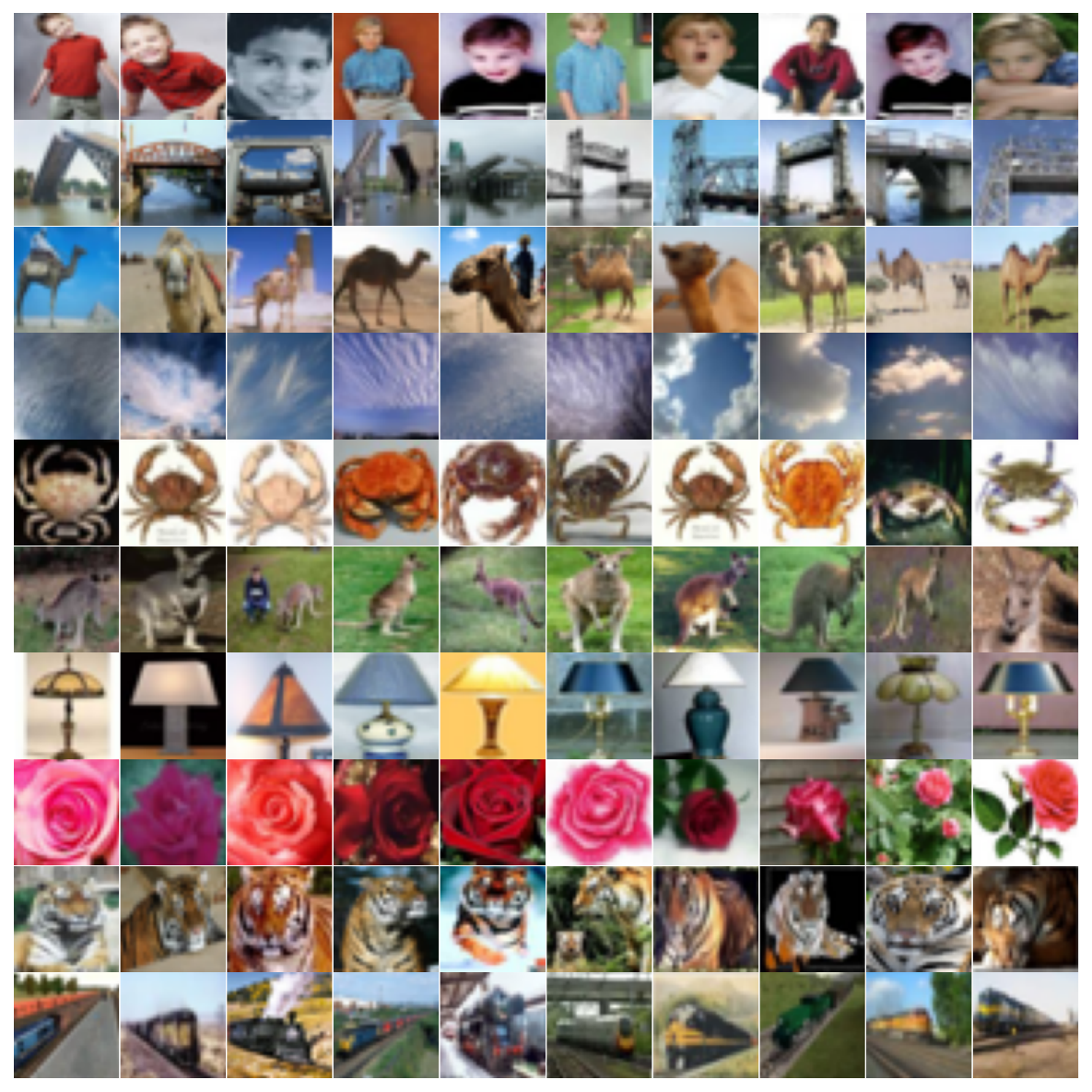}
}
\hfill
\subfigure[Challenging images from CIFAR-100]
{
\includegraphics[width=0.45\columnwidth]{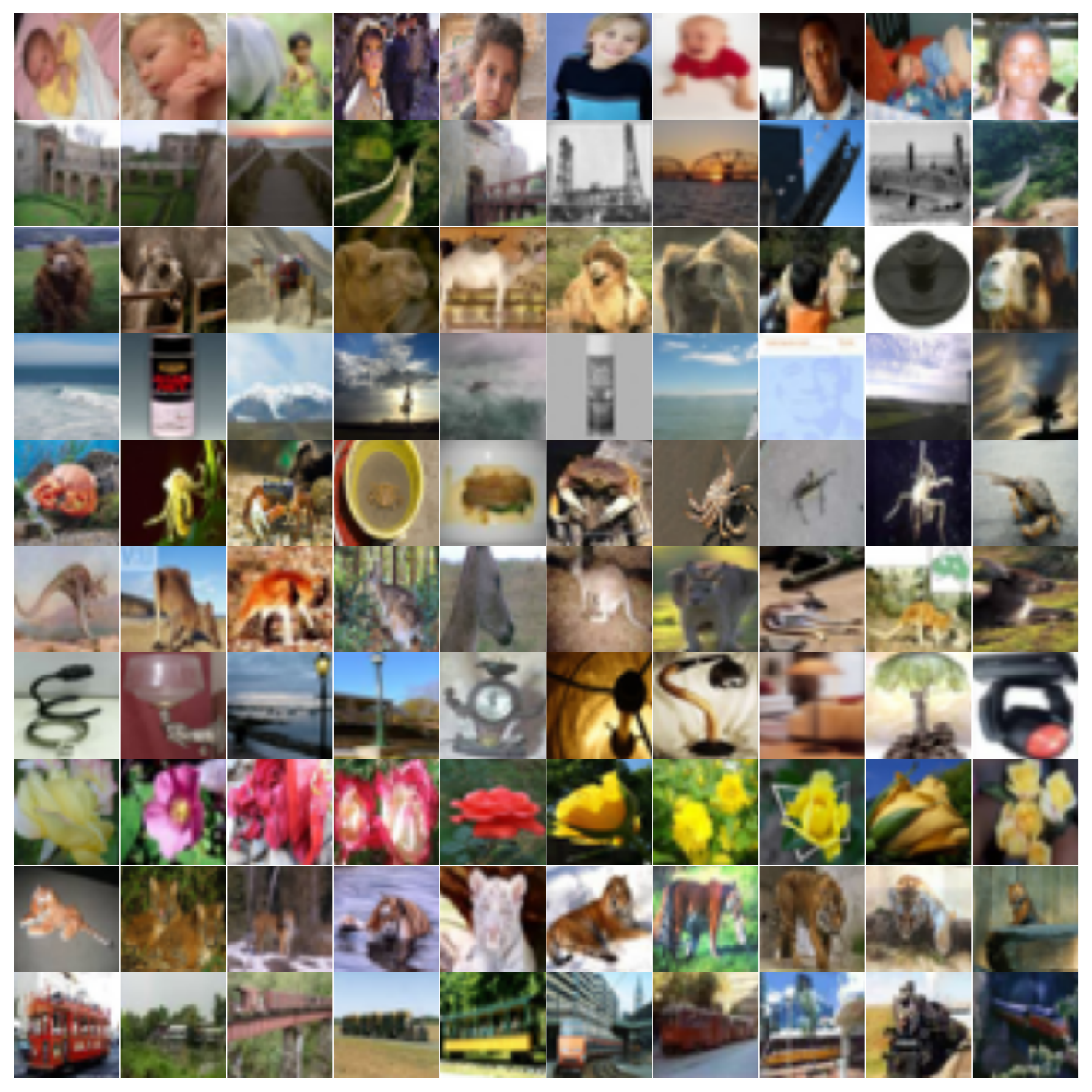}
}
\caption{Class-wise easy and challenging images for the 10 classes (\texttt{airplane, car, bird, cat, deer, dog, frog, horse, ship, truck}) in CIFAR-10 and for a subset of 10 classes (\texttt{boy, bridge, camel, cloud, crab, kangaroo, lamp, rose, tiger, train}) from CIFAR-100. Similar to the results in Fig.~\ref{fig:easy_difficult_examples}, easy images (a,c) are more canonical images associated with the class labels whereas challenging images (b,d) are images that are confused between two or more classes in the dataset.}
\label{fig:easy_difficult_examples_all}
\end{figure*}

Similar to Fig~\ref{fig:easy_difficult_examples} in Sec.~\ref{sec:visualization} of the main paper, we visualize easy and challenging examples in Fig.~\ref{fig:easy_difficult_examples_all} for CIFAR-10 and subset of classes from CIFAR-100. 
As observed the easy images (the ones that get high scores in our approach) are more canonical images of the class labels whereas the challenging ones are images that can potentially be assigned another class in the same dataset or are mislabeled in the dataset.
The clear distinctions between these images show that our concept-based score aligns well with human intuition on the difficulty of the samples.

\input{tables/affectnet}
\input{tables/bloodmnist}

\subsection{Concept-based CS for emotion recognition and biomedical image recognition}
\label{app:cs_for_other_tasks}
To validate the effectiveness of our concept-based coreset selection method beyond object recognition tasks, we apply our concept-based CS approach to the task of emotion recognition and biomedical image recognition. 

For emotion recognition, we use the Affectnet dataset~\cite{mollahosseini2017affectnet} for our experiments. 
AffectNet is a large-scale facial expression dataset designed for training and evaluating affective computing models~\cite{wang2022systematic}. 
It contains facial images collected from the internet using web search queries for emotion-related keywords in multiple languages. 
Each image is manually annotated for eight discrete emotion categories: \texttt{neutral, happiness, sadness, surprise, fear, disgust, anger, contempt}. For our experiments, we utilize an openly available version of this dataset~\footnote{https://www.kaggle.com/datasets/noamsegal/affectnet-training-data}, containing roughly $16000$ training and $14000$ testing samples. 

According to our approach we first use LLaVA to extract concepts for the 8 emotion classes, using the following prompt, \textit{``What are the facial features that distinguish {emotion class name} from other emotion types. Focus on changes in eyes, nose, lips, eyebrows, mouth. Give the output separated by commas in a line.''}. 
We get $5-10$ distinctive facial feature concepts for every emotion, for instance for emotion class \textit{happy}, we get the following concepts, \textit{``wide open eyes'', ``sparking eyes'', ``smiling lips'', ``open mouth'', ``raised eyebrows'', ``flushed cheeks'', ``teeth barred''}. We finally select $k=5$ discriminative concepts from this list. 

To test coreset performance, we use the EfficientNet model~\cite{tan2019efficientnet} and report F1 scores for our coresets in Table~\ref{Table:affectnet-results}. When compared against randomly selected coresets for the various pruning ratios, coresets selected via our concept-based approach achieve better performance.

For biomedical image recognition, we use the BloodMNIST \cite{acevedo2020dataset} dataset from the MedMNIST \cite{medmnistv1,medmnistv2} which comprises of images of normal blood cells, captured from individuals without infection, hematologic or oncologic disease and free of any pharmacologic treatment at the moment of blood collection. 
It consists of a total of $17,092$ images and is organized into 8 classes (\texttt{basophil,eosinophil,erythroblast,immature granulocytes(myelocytes, metamyelocytes and promyelocytes),lymphocyte,monocyte,neutrophil,platelet}).

For this dataset, we first extract concepts for the 8 blood cell types via GPT using the following prompt, \textit{``What are the features that can distinguish {blood cell class name} from rest of the blood cell types on their size, shape, nucleus appearance, and the presence of granules in their cytoplasm''}.
We obtain 10 concepts for every blood cell type, for instance, for \textit{platelets}, we get the following concepts, \textit{``Smallest blood component'', ``No nucleus'', ``Granules present'', ``Irregular shape'', ``Cytoplasmic fragments'', ``Variable granule distribution'', ``Oval to round shape'', ``Small dense granules'', ``Lacks chromatin'', ``Compact cytoplasmic body''}. 

To test the coreset performance, we use a ResNet-18 model and report accuracy of our coresets in Table~\ref{Table:bloodmnist-results}. 
Similar to other results, our method achieves better performance than randomly selected coresets for higher pruning rates and is competitive at lower pruning ratios.
This is attributes to the difficulty of calculating concept similarity is the representation space of the CLIP model which is potentially unaware of the terminology used in the medical domain. While replacing CLIP with a VLM that is trained on medical domain can boost the performance of our method, our results highlight that even with access to such a model our approach is able to find better coresets than random subsets. 

Our results on these two tasks highlight the versatility of our approach for model-agnostic approach to coreset selection, which is able to find coreset without requiring training the the downstream models on the entire dataset even once.

\input{tables/adaptive_cs}
\subsection{Evaluation on adaptive subset selection}
\label{app:adaptive_cs}
In this section, we present a comparison of using our concept-based method for the problem of adaptive subset selection \cite{killamsetty2021grad,killamsetty2023milo,tukan2023provable} (See App.~\ref{App:additional_rw} for a description of the problem). 
To test our approach, we followed the experimental setup of \cite{tukan2023provable}, of training a ResNet-18 model on CIFAR-10/100 for $300$ epochs and changing the subset every $20$ epochs. 
Unlike works in the line of adaptive subset selection which focus on selecting subsets that do not overlap or train models on easy subsets first before moving to difficult ones, we simply used our CCS-based selection approach \cite{zheng2022coverage} to identify new subsets every $20$ epochs for this task. 
Due to the randomness in CCS (Alg. \ref{alg:ccs}), samples being selected to form the coreset changes.

In our table~\ref{Table:adaptive_coreset_performance}, we compare our method to previous methods and with an adaptive random selection method (where a random subset of data is selected every $20$ epochs), which has been suggested by \cite{killamsetty2023milo} as a strong baseline for this line of work. 
We observe that our method without any modifications achieves comparable performance to existing adaptive subset selection methods showing its effectiveness for this problem as well. 
Moreover, sampling new subsets via CCS (Alg. \ref{alg:ccs}) is as efficient as sampling a random subset; thus our approach is also an efficient solution for this problem. 
While we believe that incorporating better subset sampling techniques as suggested by \cite{killamsetty2023milo} may boost the performance of our method on this problem, CCS-based sampling is already quite effective.

\subsection{Transferability of coresets}
\label{app:transferability_of_coresets}
In this section, we present the performance of three additional downstream model architectures after training on the coresets found by our approach. 
Similar to the results in Table~\ref{Table:transferability_of_coresets}, our results in Table~\ref{Table:transferability_of_coresets_app} show that coresets found by our approach in both standard and label-free setting achieve performance better than the random subsets on these three architectures as well. 
We note that the performance of the ViT model is worse than the performance of other model architectures with and without pruning (ViT-B-16 achieves an accuracy of $\approx 62\%$ compared to $\approx 78\%$ with ResNet-50 on full Imagenet dataset) due to the ViT models being data hungry in nature~\cite{dosovitskiy2020image} and the fact that we used standard SGD-based training (similar to that used for training other models in the paper).
We believe that using other training methods for ViTs as suggested in \cite{touvron2021training} could produce better performing ViT models.

Nonetheless, better performance than random subsets across a variety of downstream model architectures highlights the effectiveness of our approach at finding coresets without the knowledge of the architecture or training dynamics of the downstream models.

\input{tables/transferability-coreset-tm-app}

\subsection{Algorithm for stratified sampling using CCS \cite{zheng2022coverage}}
\label{app:CCS_sampling}
Here we present the algorithm for sampling the training examples to form the coreset based on the coverage-based selection methodology proposed by \cite{zheng2022coverage}.
A crucial component of the algorithm is the cutoff rate $\beta$ which controls how many challenging samples should be removed from consideration when selecting the coreset. 
This is done to eliminate misclassified samples from the dataset since they can hurt the performance of the model trained on coreset, especially at high pruning rates.
Previous works \cite{zheng2022coverage,zheng2024elfs} ablate the values of this cutoff ratio by training the downstream model on a range of values. 
In our work, we simply use the values proposed by the previous works and find that they work well for our score as well. 
The cutoff rates $\beta$ for different pruning rates $\alpha$ are as follows ($\alpha$, $\beta$).
For CIFAR-10: (30\%, 0), (50\%, 0), (70\%, 10\%), (90\%, 30\%), 
for CIFAR-100: (30\%, 10\%), (50\%, 20\%), (70\%, 20\%), (90\%, 50\%), 
for Imagenet: (30\%, 0), (50\%, 10\%), (70\%, 20\%), (90\%, 30\%).
We used CCS for label-free CS as well and the cutoff rates used were
for CIFAR-10: (30\%, 0), (50\%, 0), (70\%, 20\%), (90\%, 40\%), 
for CIFAR-100: (30\%, 0), (50\%, 20\%), (70\%, 40\%), (90\%, 50\%), 
for Imagenet: (30\%, 0), (50\%, 10\%), (70\%, 20\%), (90\%, 30\%).
% \AM{TM, can you fill the numbers for Imagenet?}

\begin{algorithm}[t] 
\caption{Coverage-centric Coreset Selection (CCS) \cite{zheng2022coverage}} 
\label{alg:ccs}
\textbf{Input}: Dataset with difficulty scores: $\mathbb{D} = \{(x,y,s)\}_{i=1}^n$, pruning ratio: $\alpha$, cutoff rate: $\beta$, number of bins: $b$. \\
\textbf{Output}: Coreset: $\mathcal{S}$
\begin{algorithmic}
\STATE{\# Prune hardest examples}
\STATE{$\mathbb{D}' \leftarrow \mathbb{D} \; \symbol{92} \; \{{\floor*{n \times \beta} \; \mathrm{hardest \; examples}}\}$}
\STATE{$A_1, A_2, \cdots, A_b \leftarrow$ Split scores in $\mathbb{D}'$ into $b$ bins.}
\STATE{$\mathcal{B} \leftarrow \{ B_i: B_i\; \mathrm{ consists \; of \; samples \; with \; scores \; in \; } A_i \; for \; i = 1, \cdots, b\}$.}
\STATE{\# Define the size of the coreset}
\STATE{$m \leftarrow n \times \alpha $.}
\STATE{}
\WHILE{$\mathcal{B} \neq \varnothing$}
    \STATE{\# Select the bin with the fewest examples}
    \STATE{$B_{min} \leftarrow \arg \min_{B \in \mathcal{B}} |B|$}.
    \STATE{\# Compute the budgets for this bin}
    \STATE{$m_{B} \leftarrow \min\{|B|, \floor*{\frac{m}{|\mathcal{B}|}}\}$}.
    \STATE{$\mathcal{S}_B \leftarrow $ randomly sample $m_B$ samples from $B_{min}$.}
    \STATE{$\mathcal{C} \leftarrow \mathcal{C} \bigcup \mathcal{S}_B.$}
    \STATE{$\mathcal{B} \leftarrow \mathcal{B} \; \symbol{92} \; \{B_{min}\}.$} 
    \STATE{$m \leftarrow m - m_{B}.$}
\ENDWHILE
\STATE{return $\mathcal{C}$.}
\end{algorithmic}
\end{algorithm}

\subsection{Experimental details}
For generating the importance score we pre-compute the concept similarity scores for the entire dataset and then train the concept-bottleneck layer (in block 2 of Fig.~\ref{fig:overview}) for $100$ epochs across all experiments. 
This training only requires 800 seconds for Imagenet which is significantly more efficient than training the ResNet-34 model on Imagenet (requires roughly 8 hours on two A-100 GPUs).
The accuracies of the models trained on the entire training set are $95.44$\% and $78.74$\% for ResNet(RN)-18 on CIFAR-10/100 and $72.4$\% for RN-18, $75$\% for RN-34, and $78.4$\% for RN-50 on Imagenet. 

After the coresets are selected, we use the setting and code from \cite{zheng2022coverage} for training a ResNet-18 model for $40000$ iterations with a batch size of $256$ on the coresets for all pruning ratios for CIFAR-10/CIFAR-100.
For Imagenet, we train ResNet-18, ResNet-34, and ResNet-50 models for $100$ epochs on the coresets identified by our method using the training code based on FFCV \cite{leclerc2023ffcv}.

The performance of the label-free CS is dependent on the quality of the pseudo-labels. 
Compared to the clustering-based approach used by ELFS \cite{zheng2024elfs}, our approach of using the zero-shot classification ability of CLIP models yields significantly better pseudo-label quality along with being simpler and more efficient to compute. 
Specifically, for CIFAR-10/100, pseudo-labels of the training set are computed using the CLIP L-14 model trained on the DataComp-1B dataset \cite{ilharco_gabriel_2021_5143773} yields an accuracy of $98.52$\% and $87.28$\% whereas for Imagenet it achieves an accuracy of $79.47$\%
% 61.7\% \AM{72.89, 79.47} 
which are better than the best pseudo-label accuracy obtained by the clustering approach in ELFS ($92.5$\% and $66.3$\% on CIFAR-10/100 and $58.8$\% on Imagenet).

For training the concept bottleneck layer we minimized the cross entropy loss using SGD with a learning rate of 1E-3, momentum of $0.9$ and a weight decay of 5E-4 for $100$ epochs.

% \AM{TODO:Additional details of training the bottleneck layers.}

 %misratios
% write_stratified_dataset () {

    % if [ $5 == 0.1 ]
    % then 
    %     misratio=0.3
    % elif [ $5 == 0.3 ]
    % then
    %     misratio=0.2
    % elif [ $5 == 0.5 ]
    % then
    %     misratio=0.1
    % else
    %     misratio=0.
    % fi
% }

%% file: tables/affectnet.tex
\begin{table}[tb]
  \caption{\small{
  % Performance of models with different architectures on our coresets for Imagenet. 
  Performance of concept-based coreset selection for emotion recognition task on Affectnet dataset. Coresets selected by our approach results in superior performance (higher F1 score) compared to Random for standard CS.
   }}
  \label{Table:affectnet-results}
  \centering
  \small
  \resizebox{0.75\linewidth}{!}{
    \begin{tabular}{crcccc}
    \toprule[1.25pt]
    \multirow{2}[1]{*}{\makecell{\textbf{Model}\\ \textbf{Arch.}}} & \multirow{2}[1]{*}{\textbf{Method}} &  \multicolumn{4}{c}{\makecell{\textbf{Pruning Rates}}}\\ 
    \cmidrule(ll){3-6}
    & & $30\%$ & $50\%$ & $70\%$ & $90\%$ \\
    \midrule
    \multirow{2}[2]{*}{\textbf{EfficientNet}} & \textbf{Random} & $0.607_{\scriptscriptstyle \pm0.006}$ & $0.573_{\scriptscriptstyle \pm0.025}$ & $0.517_{\scriptscriptstyle \pm0.006}$ & $0.347_{\scriptscriptstyle \pm0.055}$ \\
    \cmidrule(lr){2-6}
    & \textbf{Ours} & $0.603_{\scriptscriptstyle \pm0.006}$ & $0.577_{\scriptscriptstyle \pm0.021}$ & $0.537_{\scriptscriptstyle \pm0.015}$ & $0.450_{\scriptscriptstyle \pm0.035}$  \\
    \bottomrule[1.25pt]
    \end{tabular}
    }
     \vspace{-0.3cm}
\end{table}

%% file: tables/bloodmnist.tex
\begin{table}[tb]
  \caption{\small{
  Performance of concept-based coreset selection for biomedical image recognition task on BloodMNIST dataset. Coresets selected by our approach results in superior accuracy compared to Random for standard CS, highlighting our method's effectiveness across various tasks.
   }}
  \label{Table:bloodmnist-results}
  \centering
  \small
  \resizebox{0.75\linewidth}{!}{
    \begin{tabular}{crcccc}
    \toprule[1.25pt]
    \multirow{2}[1]{*}{\makecell{\textbf{Model}\\ \textbf{Arch.}}} & \multirow{2}[1]{*}{\textbf{Method}} &  \multicolumn{4}{c}{\makecell{\textbf{Pruning Rates}}}\\ 
    \cmidrule(ll){3-6}
    & & $30\%$ & $50\%$ & $70\%$ & $90\%$ \\
    \midrule
    \multirow{2}[2]{*}{\textbf{ResNet-18}} & \textbf{Random} & $94.99_{\scriptscriptstyle \pm0.27}$ & $94.79_{\scriptscriptstyle \pm0.18}$ & $91.71_{\scriptscriptstyle \pm0.29}$ & $86.50_{\scriptscriptstyle \pm0.66}$ \\
    \cmidrule(lr){2-6}
    & \textbf{Ours} & $94.64_{\scriptscriptstyle \pm0.60}$ & $94.06_{\scriptscriptstyle \pm0.10}$ & $92.26_{\scriptscriptstyle \pm0.19}$ & $87.44_{\scriptscriptstyle \pm1.46}$  \\
    \bottomrule[1.25pt]
    \end{tabular}
    }
     \vspace{-0.3cm}
\end{table}

%% file: tables/adaptive_cs.tex
\begin{table*}[tb]
  \caption{\small{Adaptive coreset selection (Results for RBFNN and GradMatchPB are taken from tables 1 and 3 of Tukan et. al.)
  }}
  \label{Table:adaptive_coreset_performance}
  \centering
  \small
  \resizebox{0.75\textwidth}{!}{
    \begin{tabular}{rcccc}
    \toprule[1.25pt]
   \multirow{3}[4]{*}{\textbf{Method}} &  \multicolumn{4}{c}{\makecell{\textbf{Datasets and Pruning Rates}}} \\ 
   \cmidrule(lr){1-4}
   &  \multicolumn{2}{c}{\makecell{\textit{CIFAR-10}}} & \multicolumn{2}{c}{\makecell{\textit{CIFAR-100}}} \\
    \cmidrule(lr){2-3} \cmidrule(lr){4-5}
   &  $70\%$ & $90\%$ & $70\%$ & $90\%$ \\
   
    \midrule
    GradMatchPB \cite{killamsetty2021grad}   & $91.89$ & $90.01$ & $72.57$ & $60.39$ \\
    RBFNN \cite{tukan2023provable} & ${94.44}$ & ${91.40}$ & $73.48$ & $64.59$  \\
    Adaptive Random  & $93.64_{\scriptscriptstyle \pm0.21}$ & $90.49_{\scriptscriptstyle \pm0.41}$ & ${71.70}_{\scriptscriptstyle \pm0.30}$ & ${61.24}_{\scriptscriptstyle \pm0.37}$ \\
    Ours (adaptive) & $92.61_{\scriptscriptstyle \pm0.08}$ & $89.95_{\scriptscriptstyle \pm0.26}$ & ${71.74}_{\scriptscriptstyle \pm0.33}$ & $63.71_{\scriptscriptstyle \pm0.48}$ \\
    
    \bottomrule[1.25pt]
    \end{tabular}
    }
    \vspace{-0.2cm}
\end{table*}

%% file: tables/transferability-coreset-tm-app.tex
% \begin{wraptable}[23]{r}{8.5cm}
\begin{table}[tb]
% \vspace{-0.6cm}
  \caption{\small{% Performance of models with different architectures on our coresets for Imagenet. 
  Superior performance of downstream models with different architectures trained on our coresets for Imagenet compared to Random for both standard (Ours) and label-free (Ours-LF) CS highlight our transferability of our coresets.
  % Results of standard (Ours) and label-free (Ours-LF) CS show that models trained on our coresets perform better than random subsets of data regardless of the architecture of the downstream model. 
  % models trained on our coreset perform better than those trained on a random subsets of data. 
   % \multirow{2}{*}{\makecell{\textbf{Needs Training} \\ \textbf{Dynamics}}
   }}
  \label{Table:transferability_of_coresets_app}
  \centering
  \small
  \resizebox{0.6\columnwidth}{!}{
    \begin{tabular}{crcccc}
    \toprule[1.25pt]
    \multirow{2}[1]{*}{\makecell{\textbf{Model}\\ \textbf{Architecture}}} & \multirow{2}[1]{*}{\textbf{Method}} &  \multicolumn{4}{c}{\makecell{\textbf{Pruning Rates}}}\\ 
    \cmidrule(ll){3-6}
    & & $30\%$ & $50\%$ & $70\%$ & $90\%$ \\

    \midrule
    \multirow{3}[2]{*}{\textbf{MobileNet}} & \textbf{Random} & $62.68_{\scriptscriptstyle \pm0.09}$ & $62.49_{\scriptscriptstyle \pm0.15}$ & $61.53_{\scriptscriptstyle \pm0.29}$ & $53.80_{\scriptscriptstyle \pm0.17}$ \\
    \cmidrule(lr){2-6}
    & \textbf{Ours-LF} & $61.60_{\scriptscriptstyle \pm0.18}$ & $61.97_{\scriptscriptstyle \pm0.08}$ & $61.73_{\scriptscriptstyle \pm0.25}$ & $54.39_{\scriptscriptstyle \pm0.66}$\\
    & \textbf{Ours} & $61.36_{\scriptscriptstyle \pm0.15}$ & $62.32_{\scriptscriptstyle \pm0.22}$ & $62.46_{\scriptscriptstyle \pm0.26}$ & $55.57_{\scriptscriptstyle \pm0.13}$  \\

    \midrule
    \multirow{3}[2]{*}{\textbf{RN-34}}& \textbf{Random} & $73.37_{\scriptscriptstyle \pm0.08}$ & $71.71_{\scriptscriptstyle \pm0.10}$ & $67.85_{\scriptscriptstyle \pm0.04}$ & $51.29_{\scriptscriptstyle \pm0.20}$ \\
    \cmidrule(lr){2-6}
    & \textbf{Ours-LF} & $73.61_{\scriptscriptstyle \pm0.08}$ & $71.99_{\scriptscriptstyle \pm0.05}$ & $68.42_{\scriptscriptstyle \pm0.21}$ & $53.52_{\scriptscriptstyle \pm0.06}$\\
    & \textbf{Ours} & $73.39_{\scriptscriptstyle \pm0.12}$ & $72.34_{\scriptscriptstyle \pm0.13}$ & $69.44_{\scriptscriptstyle \pm0.17}$ & $55.92_{\scriptscriptstyle \pm0.02}$ \\
    \midrule
    
    \multirow{3}[2]{*}{\textbf{ViT-B-16}} & \textbf{Random} & $59.09_{\scriptscriptstyle \pm0.49}$ & $51.65_{\scriptscriptstyle \pm 0.46}$ & $40.19_{\scriptscriptstyle \pm0.13}$ & $22.13_{\scriptscriptstyle \pm0.16}$ \\
    \cmidrule(lr){2-6}
    & \textbf{Ours-LF} & $57.50_{\scriptscriptstyle \pm0.69}$ & $49.81_{\scriptscriptstyle \pm0.89}$ & $40.33_{\scriptscriptstyle \pm0.86}$ & $23.08_{\scriptscriptstyle \pm0.27}$\\
    & \textbf{Ours} & $57.62_{\scriptscriptstyle \pm0.56}$ & $52.67_{\scriptscriptstyle \pm0.49}$ & $42.15_{\scriptscriptstyle \pm 0.81}$ & $24.43_{\scriptscriptstyle \pm 0.28}$ \\
    \bottomrule[1.25pt]
    \end{tabular}
    }
\end{table}
% \end{wraptable}